\pdfoutput=1

\documentclass[11pt]{article}

\usepackage[final]{acl}

\usepackage[export]{adjustbox} 
\usepackage{times}
\usepackage{latexsym}
\usepackage{graphicx}
\usepackage{amsmath}
\usepackage{subcaption}
\usepackage{arydshln}
\usepackage{multirow}
\usepackage{booktabs}
\usepackage{float}
\usepackage{pgfplots}
\pgfplotsset{compat=1.17}
\usepackage{lipsum} 

\usepackage{xcolor}

\definecolor{darkgreen}{rgb}{0.0, 0.5, 0.0} 

\definecolor{blue}{RGB}{68, 114,196}
\definecolor{orange}{RGB}{237, 125,49}

\definecolor{battleshipgrey}{rgb}{0.3, 0.3, 0.3}
\definecolor{brilliantrose}{rgb}{1.0, 0.33, 0.64}
\definecolor{americanrose}{rgb}{1.0, 0.01, 0.24}
\definecolor{jweigreen}{rgb}{0,0.45,0.24}
\definecolor{bluegray}{rgb}{0.1, 0.1, 0.4}
\definecolor{ao(english)}{rgb}{0.0, 0.5, 0.0}
\definecolor{blanchedalmond}{rgb}{1.0, 0.92, 0.8}
\definecolor{atomictangerine}{rgb}{1.0, 0.6, 0.4}
\definecolor{chocolate(web)}{rgb}{0.82, 0.41, 0.12}
\definecolor{bananayellow}{rgb}{1.0, 0.88, 0.21}
\definecolor{goldenbrown}{rgb}{0.6, 0.4, 0.08}
\definecolor{aliceblue}{rgb}{0.94, 0.97, 1.0}
\definecolor{beige}{rgb}{0.96, 0.96, 0.86}
\definecolor{babyblue}{rgb}{0.54, 0.81, 0.94}
\definecolor{camel}{rgb}{0.76, 0.6, 0.42}
\definecolor{cinnamon}{rgb}{0.82, 0.41, 0.12}
\definecolor{deepskyblue}{rgb}{0.0, 0.75, 1.0}
\definecolor{frenchblue}{rgb}{0.0, 0.45, 0.73}
\definecolor{classicrose}{rgb}{0.98, 0.8, 0.91}
\definecolor{frenchrose}{rgb}{0.96, 0.29, 0.54}
\definecolor{frenchlilac}{rgb}{0.53, 0.38, 0.56}
\definecolor{frenchbeige}{rgb}{0.65, 0.48, 0.36}
\definecolor{metablue}{HTML}{0064E0}
\definecolor{googlegreen}{HTML}{253d7b}
\definecolor{mistralorange}{HTML}{f37004}
\definecolor{openaigreen}{HTML}{10a37f}
\definecolor{mt_purple}{HTML}{C8A1C8}
\definecolor{example_src}{HTML}{F6A503}
\definecolor{example_tgt}{HTML}{A5292A}

\definecolor{llama2blue}{RGB}{173, 216, 230}
\definecolor{tower_mono_blue}{RGB}{95, 158, 160}
\definecolor{tower_green}{RGB}{86, 163, 74}
\definecolor{tower_inst_green}{RGB}{58, 125, 48}
\definecolor{random_purple}{RGB}{128, 0, 128}

\usepackage[most]{tcolorbox}
\newtcbox{\hlexampleone}{on line, rounded corners, box align=base, colback=example_src!18,colframe=white,size=fbox,arc=3pt, before upper=\strut, top=-2pt, bottom=-4pt, left=-2pt, right=-2pt, boxrule=0pt}
\newtcbox{\hlexampleonetgt}{on line, rounded corners, box align=base, colback=example_tgt!18,colframe=white,size=fbox,arc=3pt, before upper=\strut, top=-2pt, bottom=-4pt, left=-2pt, right=-2pt, boxrule=0pt}

\newtcbox{\hlsrc}{on line, rounded corners, box align=base, colback=tower_green!30,colframe=white,size=fbox,arc=3pt, before upper=\strut, top=-2pt, bottom=-4pt, left=-2pt, right=-2pt, boxrule=0pt}
\newtcbox{\hlmt}{on line, rounded corners, box align=base, colback=mt_purple!50,colframe=white,size=fbox,arc=3pt, before upper=\strut, top=-2pt, bottom=-4pt, left=-2pt, right=-2pt, boxrule=0pt}

\usepackage{soul}

\setul{0.5ex}{0.3ex}
\setulcolor{bananayellow}

\makeatletter
\def\adl@drawiv#1#2#3{%
        \hskip.5\tabcolsep
        \xleaders#3{#2.5\@tempdimb #1{1}#2.5\@tempdimb}%
                #2\z@ plus1fil minus1fil\relax
        \hskip.5\tabcolsep}
\newcommand{\cdashlinelr}[1]{%
  \noalign{\vskip 2pt
           \global\let\@dashdrawstore\adl@draw
           \global\let\adl@draw\adl@drawiv}
  \cdashline{#1}[.4pt/2pt]
  \noalign{\global\let\adl@draw\@dashdrawstore
           \vskip 2pt}}
\makeatother

\usepackage[T1]{fontenc}

\usepackage[utf8]{inputenc}

\usepackage{microtype}

\usepackage{inconsolata}
\usepackage{amsfonts} 
\usepackage{bm}
\usepackage{pifont}
\usepackage{xcolor}
\usepackage{tikz}
\usepackage{amssymb}
\usetikzlibrary{shapes, arrows, positioning}

\title{Analyzing Context Contributions in LLM-based Machine Translation}

\author{
    Emmanouil Zaranis\textsuperscript{\normalfont 1,2}, 
    Nuno M. Guerreiro\textsuperscript{\normalfont1,2,3,4}, 
    André F. T. Martins\textsuperscript{\normalfont1,2,3} \\
    \textsuperscript{1}Instituto de Telecomunicações,
    \textsuperscript{2}Instituto Superior Técnico,
    \textsuperscript{3}Unbabel,
    \textsuperscript{4}MICS\\
    \texttt{emmanouil.zaranis@tecnico.ulisboa.pt}
}

\begin{document}
\maketitle
\begin{abstract}
Large language models (LLMs) have achieved state-of-the-art performance in machine translation (MT) and demonstrated the ability to leverage in-context learning through few-shot examples. However, the mechanisms by which LLMs use different parts of the input context remain largely unexplored. In this work, we provide a comprehensive analysis of context utilization in MT, studying how LLMs use various context parts, such as few-shot examples and the source text, when generating translations. 
We highlight several key findings: (1) the source part of few-shot examples appears to contribute more than its corresponding targets, irrespective of translation direction; (2) finetuning LLMs with parallel data alters the contribution patterns of different context parts; and (3) there is a positional bias where earlier few-shot examples have higher contributions to the translated sequence. Finally, we demonstrate that inspecting anomalous context contributions can potentially uncover pathological translations, such as hallucinations. Our findings shed light on the internal workings of LLM-based MT which go beyond those known for standard encoder-decoder MT models. 
\end{abstract}

\section{Introduction}
Large language models (LLMs) have reached state-of-the-art performance in machine translation (MT) and are making significant strides toward becoming the \textit{de facto} solution for neural MT~\cite{kocmi-etal-2023-findings,alves-2024-tower-llm}. Compared to the classical standard approach using encoder-decoder models~\citep{bahdanau2016neural, vaswani2017attention}, LLMs are typically decoder-only models parameterized by billions of parameters. Remarkably, LLMs have demonstrated the ability to perform translation tasks without being explicitly trained for them, instead leveraging in-context learning (ICL) through demonstrations of the task \cite{zhang2022opt,agrawal-etal-2023-example_selection_MT,henry_gpt-mt-2023,alves-etal-2023-steering, garcia2023unreasonable}. 
Yet, there is a gap in the literature on understanding the internal workings of LLM-based MT. Previous interpretability research on MT has been limited to traditional, specialized encoder-decoder models \cite{ding-etal-2017-visualizing-and-understanding-nmt,ferrando-etal-2022-towards_opening_the_black_box-alti+,ferrando-etal-2022-measuring_the_mixing-alti,voita-etal-2021-analyzing_source_target_contributions, sarti_iclr_2024_quantifying_plausibility_context_reliance_MT_recore, mohammed-niculae-acl_2024-measuring_context_utilization_MT}, and while substantial work has investigated ICL in other tasks, such as classification \cite{min-etal-emnlp-2022-rethinking_the_role_of_demonstrations,lu-etal-2022-fantastically_ordered_prompts_order_sensitivity,yoo-etal-emnlp-2022-ground-truth-labels-matter,wang-etal-2023-label-words} and question answering \cite{liu-etal-2022-acl_what_makes_good_in_context_examples,liu_2023_arxiv_lost_in_the_middle,si_acl_2023_measuring_inductive_biases_icl,wei_2023_llms_do_icl_differently}, the mechanisms by which LLMs leverage \textit{parts} of context in MT remain largely unexplored.

In this work, we aim to fill this research gap by contributing towards a better understanding of how LLMs utilize different parts of the provided context~(\textit{e.g.}, few-shot examples, the source text, or previously generated target tokens) in MT. While previous work conducted on understanding the impact of context in MT largely focuses on performing modifications on the LLM input and measuring performance drop \citep{Zhu_2023_MultilingualMT_Empirical_Analysis,raunak-etal-acl-2023-dissecting-icl-in-gpts}, we take instead an attribution-based approach~\cite{ferrando-etal-2022-towards_opening_the_black_box-alti+}, tracking the input tokens’ relevance in all parts of the context---this allows us to estimate how different parts of context contribute to the generated translations, providing a more fine-grained analysis of context utilization.

\begin{figure*}[t]
    \centering
    \input{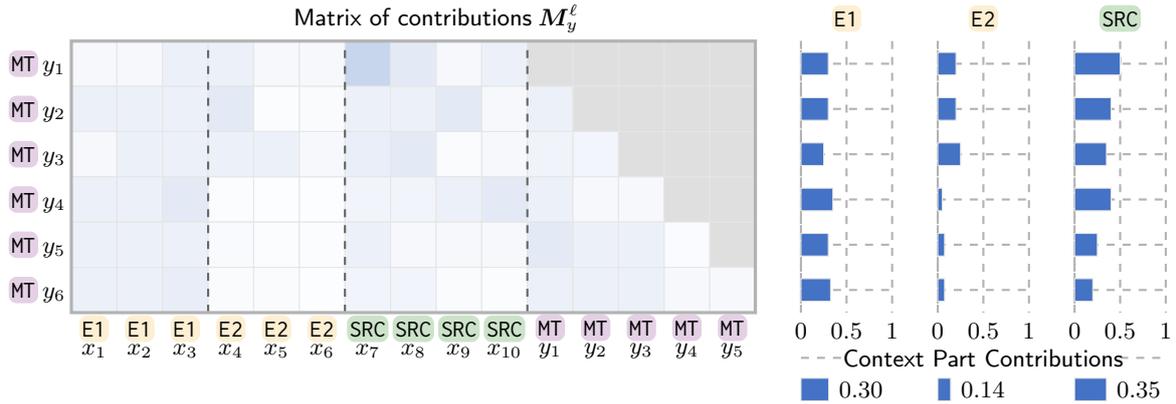}    
    \caption{Illustration of \textit{synthetic} part-level \textit{total} contributions computation given 2 examples as context. From the token-to-token level contribution matrix $\bm{M}_y^\ell$, we compute the total contribution of each input part to each generated token, by summing the corresponding token-level contributions. Subsequently,  we compute the part-level total contribution of each input part to the translated sequence, by averaging over the generated tokens.} 
\label{fig:alti_contr_total}
\end{figure*}

We study several key aspects of context utilization in MT using general purpose LLaMA-2 models \cite{touvron-2023-llama2} and \textsc{Tower} models \cite{alves-2024-tower-llm}---a suite of models specifically adapted for translation tasks. First, we investigate how different input parts contribute to the translated sequence. Next, we explore whether the provided few-shot examples contribute equally to the translated sequence. We also analyze if undergoing adaptation via continuous pretraining~\citep{gupta2023continual, yıldız2024investigating, alves-2024-tower-llm} on relevant multilingual and parallel data leads to a change in these contribution patterns. Moreover, to further understand the translation dynamics, we examine how context contributions vary at different stages of the generation process. Finally, we also assess whether anomalous context contributions can uncover catastrophic translations, such as hallucinations~\cite{dale-etal-acl-2023-detecting-and-mitigating-hallucinations}. 

Our analysis reveals several key insights on context utilization by LLMs for translation, including:
\begin{itemize}
    \item Irrespective of the translation direction, the source of each few-shot example contributes more than its corresponding target;
    \item  The examined models exhibit a positional bias---earlier few-shot examples tend to have higher contributions to the translated sequence. Additionally, the bias is maintained across different generation stages;
    \item Training on task-specific data reduces the influence of few-shot examples and consequently shrinks the positional bias observed;
    \item  Low source contributions can potentially uncover pathological translations.
\end{itemize}

We release all our code, and make our results available across all tested models and languages.\footnote{\url{https://github.com/deep-spin/interp_llm}}

\section{Problem Formulation}
\label{sec:problem_formulation}
In this section, we introduce ICL and describe how we employ the ALTI method \cite{ferrando-etal-2022-towards_opening_the_black_box-alti+} to measure the contribution of each input \textit{part} in the context to the translated sequence.

\subsection{In-Context Learning (ICL)}
ICL is a paradigm where LLMs "learn" to solve new tasks at inference time by being provided with a few task demonstrations as part of the input prompt, without requiring any updates to their parameters or fine-tuning~\citep{brown2020language, agrawal-etal-2023-example_selection_MT, henry_gpt-mt-2023}. More broadly, for MT, few-shot examples can also be used for inference time adaptation, \textit{e.g.} to different domains, terminology, or other elements of translation, guiding the model to produce outputs that are more suitable for the given context~\citep{alves-etal-2023-steering, aycock-bawden-2024-topic}.

\subsection{ALTI for autoregressive language models} 
\label{sec:ALTI_method} 

For our analysis, we choose the ALTI (Aggregation of Layer-Wise Token-to-Token Interactions) method~\cite{ferrando-etal-2022-towards_opening_the_black_box-alti+} for its simplicity and proven success in various applications. ALTI has been successfully employed for detecting hallucinations in MT \cite{dale-etal-2023-halomi_hallucination_detection_MT,guerreiro_2023_hallucinations_in_llms}, identifying toxicity in multilingual text~\cite{nllbteam2022language, costa-jussa-etal-2023-toxicity}, and explaining information flows in LLMs~\cite{ferrando_voita_2024_information_flow,tufanov_voita_2024_lm_transparency_tool}.

ALTI is an input attribution method that quantifies the mixing of information in the transformer architecture~\cite{vaswani2017attention}. It follows the modeling approach proposed by~\citet{abnar-zuidema-2020-quantifying_attention_flow-attention_rollout}, where the information flow in the model is simplified as a directed acyclic graph, with nodes representing token representations and edges representing the influence of each input token representation on the output token representation~(for each layer of the transformer). ALTI proposes using token contributions instead of raw attention weights, and computes the amount of information flowing from one node to another in different layers by summing over the different paths connecting both nodes, where each path is the result of the multiplication of every edge in the path. Formally, given an input sequence of length $S$ and an output sequence of length $T$, we compute a token-to-token contribution matrix $\bm{C}^\ell\in \mathbb{R}^{(S+T) \times (S+T)}$, where $\ell$ is the $\ell$-th layer of the model.\footnote{Note that this matrix is causal masked.} The element $c^\ell_{i,j}$ of the matrix represents the contribution of the $j$-th input token at layer $
\ell-1$ to the $i$-th output token at layer $\ell$. By multiplying the layer-wise coefficient matrices, $\bm{M}^\ell = \bm{C}^\ell \cdot \bm{C}^{\ell -1} \cdots \bm{C}^1$\, we can describe representations of intermediate layers (and final layer) as a linear combination of the model input tokens---an example of a contribution matrix is shown in Figure \ref{fig:alti_contr_total}.\footnote{For simplicity, we will consider $\bm{M}_y^\ell$ as the matrix containing the last $T$ rows of $\bm{M}^\ell$---these rows contain the contributions of the input parts to the output tokens.} This matrix can be used to interpret the model's behavior and study how different parts of the input influence generated outputs. For more details, see~\citet{ferrando-etal-2022-towards_opening_the_black_box-alti+}.

\subsection{Part-level contributions}
\label{sec:part-level_contr}
To quantify the contribution of each input part to the translated sequence, we perform a two-step aggregation process, illustrated in Figure \ref{fig:alti_contr_total}. First, we compute the total contribution of each part to each generated token by summing the corresponding token-level contributions within each part~(right hand-side of Figure~\ref{fig:alti_contr_total}). Then, we average the part-to-token contributions across the generated tokens to compute the contributions of each context part to the entire translated sequence. Similarly to \cite{ferrando-etal-2022-towards_opening_the_black_box-alti+, dale-etal-acl-2023-detecting-and-mitigating-hallucinations, dale-etal-2023-halomi_hallucination_detection_MT, guerreiro_2023_hallucinations_in_llms}, these part-level contributions are used for the analysis in the following sections.\footnote{We follow previous work and analyze the last-layer contributions.}

\section{Experimental Setup}\label{sec:experimental_setup}
We provide an overview of the
models and datasets used throughout our study, as well as important considerations on how we prompt the models.

\paragraph{Models.} We experiment with two families of models: the general-purpose \textsc{Llama-2} 7B base model \citep{touvron-2023-llama2}, and the state-of-the-art \textsc{Tower} 7B base model, which is a continued pretrained checkpoint of \textsc{Llama-2} 7B on a mixture of monolingual and parallel data~\citep{alves-2024-tower-llm}. We also experiment with \textsc{TowerInstruct} 7B, which is obtained via finetuning \textsc{Tower} on a set of instructions for translation-related tasks.\footnote{We use the following HuggingFace checkpoints: \textsc{Llama-2} (\texttt{meta-llama/Llama-2-7b-hf}), \textsc{Tower} (\texttt{Unbabel/TowerBase-7B-v0.1}), and \textsc{TowerInstruct} (\texttt{Unbabel/TowerInstruct-7B-v0.2}).}

\begin{figure*}[ht]
    \centering
    \includegraphics[width=\textwidth]{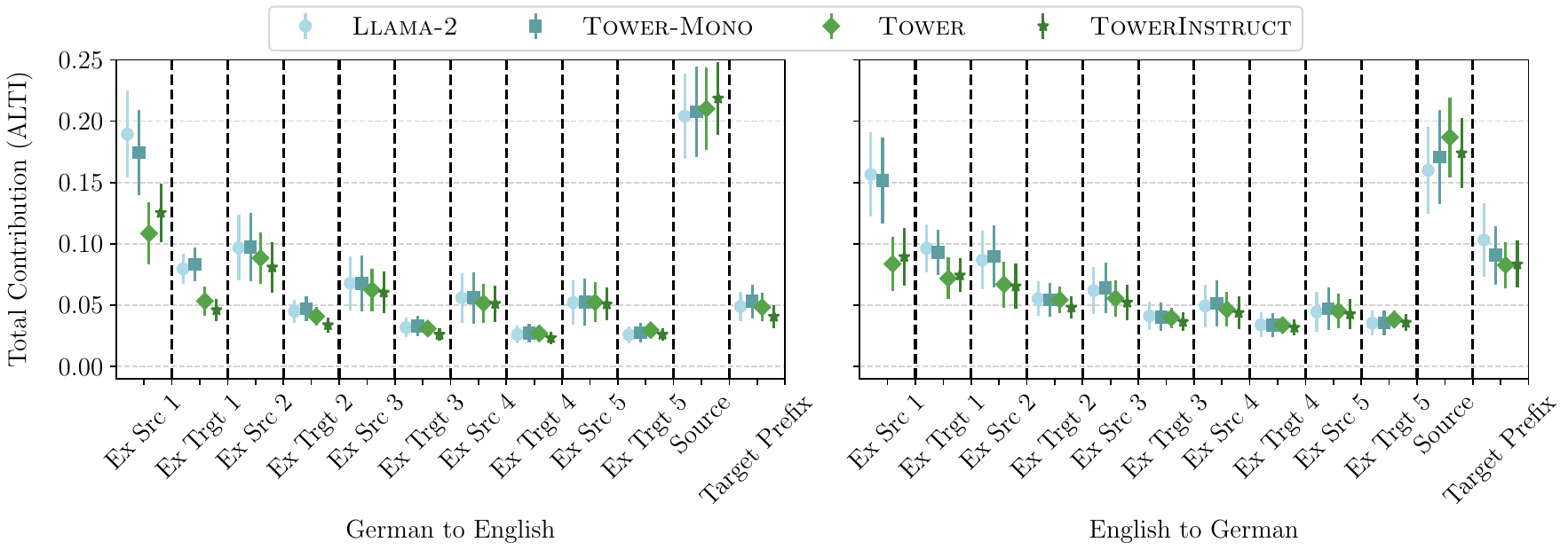}
    \captionsetup{width=\textwidth}
    \caption{Illustration of context's part-level contributions to the translated sequence, for all the examined models.}
\label{fig:pos_bias_dataset_level_de_en_rel}
\end{figure*}

\paragraph{Datasets.} We conduct our study on the publicly available WMT22 test sets, examining English to German (\texttt{en-de}),
German to English (\texttt{de-en}), English to Russian (\texttt{en-ru}) and
Russian to English (\texttt{ru-en}) language pairs.\footnote{German is the second most frequent language in \textsc{Llama-2}~\citep{touvron-2023-llama2}, just behind English, while Russian accounts for approximately 0.13\% of the training data.}

\paragraph{Few-shot setting and prompt selection.} We conduct our analysis under a 5-shot setting, using the few-shot examples provided by \citealt{henry_gpt-mt-2023}, which were selected to be high-quality examples and relevant---according to embedding similarity---to the source text. We make sure that the examples in the context are shuffled and not sorted by relevance to the source.\footnote{We include experiments with a different shuffling seed in Appendix~\ref{app:top_level_analysis}---trends in results are similar to those reported in the main text.} We use the prompt templates suggested in \citealt{zhang-icml-2023-prompting-llms-for-mt}. Additional details are provided in Appendix~\ref{app:few_shot_selection}.

\paragraph{Filtering.} Due to the high GPU memory requirements of the attribution method when applied to a 7B parameter model, we had to filter samples with large context length. We provide more details about the filtering process in Appendix \ref{app:filtering_details}.

\section{How Do Different Context Parts Contribute to the Translated Sequence?} \label{sec1:high_level_analysis_segment_contributions}
In this section, we conduct a top-level analysis by measuring and comparing the contributions of different input parts to the generated translation.

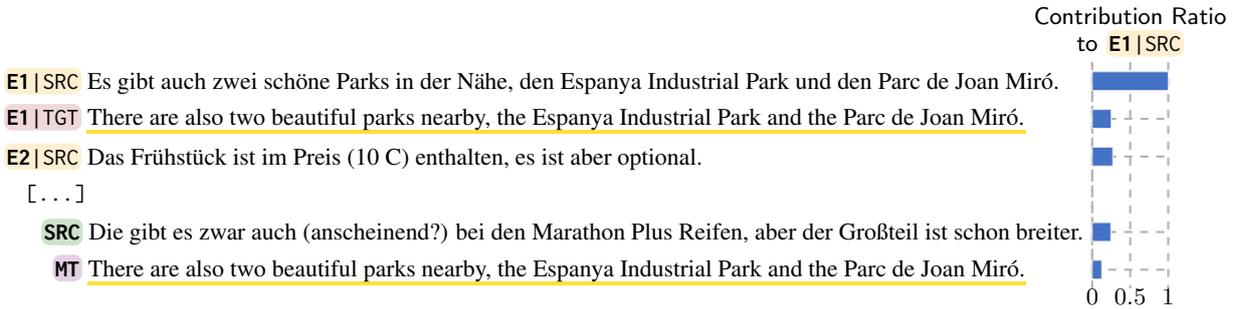
\begin{figure*}[t]
    \centering
    \begin{tikzpicture}[scale=0.5, every node/.style={font=\footnotesize, inner sep=0,outer sep=0}]
\node [] (A) at (-0.75,5.5) {\hlexampleone{\texttt{\textbf{E1}|SRC}} Es gibt auch zwei schöne Parks in der Nähe, den Espanya Industrial Park und den Parc de Joan Miró.};
\node [below=0.5cm of A.west,anchor=west] (B) {\hlexampleonetgt{\texttt{\textbf{E1}|TGT}} \ul{There are also two beautiful parks nearby, the Espanya Industrial Park and the Parc de Joan Miró.}};
\node [below=0.5cm of B.west,anchor=west] (C) {\hlexampleone{\texttt{\textbf{E2}|SRC}} Das Frühstück ist im Preis (10 C) enthalten, es ist aber optional.};
\node [below=0.5cm of C.west,anchor=west] (E) {\hspace{6pt} \texttt{[...]}};
\node [below=0.5cm of E.west,anchor=west] (F) {\hspace{14pt}\hlsrc{\texttt{\textbf{SRC}}} Die gibt es zwar auch (anscheinend?) bei den Marathon Plus Reifen,
aber der Großteil ist schon breiter.};
\node [below=0.5cm of F.west,anchor=west] (G) {\hspace{16pt} \hlmt{\texttt{\textbf{MT}}} \ul{There are also two beautiful parks nearby, the Espanya Industrial Park and the Parc de Joan Miró.}};

\draw (15,7.25) node {\textsf{Contribution Ratio}};
\draw (15,6.5) node {\textsf{to \hlexampleone{\texttt{\textbf{E1}|SRC}}}};
\draw[thick, dashed, gray!60] (14,0) -- (14,6);
\draw[thick, dashed, gray!60] (14,6) -- (14,0) node[anchor=north, black] {$0$};
\draw[thick, dashed, gray!60] (15,6) -- (15,0) node[anchor=north, black] {$0.5$};
\draw[thick, dashed, gray!60] (16,6) -- (16,0) node[anchor=north, black] {$1$};
\draw[thick, dashed, gray!60] (14,5.5) -- (16,5.5);
\filldraw[blue, draw=gray!20] (14,5.25) rectangle (16,5.75);
\draw[thick, dashed, gray!60] (14,4.5) -- (16,4.5);
\filldraw[blue, draw=gray!20] (14,4.25) rectangle (14.5,4.75);
\draw[thick, dashed, gray!60] (14,3.5) -- (16,3.5);
\filldraw[blue, draw=gray!20] (14,3.25) rectangle (14.54,3.75);
\draw[thick, dashed, gray!60] (14,1.5) -- (16,1.5);
\filldraw[blue, draw=gray!20] (14,1.25) rectangle (14.49,1.75);
\draw[thick, dashed, gray!60] (14,0.5) -- (16,0.5);
\filldraw[blue, draw=gray!20] (14,0.25) rectangle (14.25,0.75);
\end{tikzpicture}
    \caption{Example of anomalous source contributions for \textsc{Tower} which hallucinates, copying information from the first example. We show contribution ratios to \hlexampleone{\texttt{\textbf{E1}|SRC}}---$1$ being the contribution of \hlexampleone{\texttt{\textbf{E1}|SRC}}.}
    \label{fig:anomalous_mt}
\end{figure*}

\subsection{Analysis setup} 
To investigate the contribution of different prompt parts to the translated sequence, we first divide the context into the following parts: source and target side of each few-shot example, source text, and target prefix. Then, we follow the approach described in Section~\ref{sec:part-level_contr} and obtain part-level contributions that are used for analysis.

\subsection{Results} 
In Figure \ref{fig:pos_bias_dataset_level_de_en_rel}, we show, for all the examined models, the total contribution of each context part to the translated sequence for German to English and English to German language pairs.

\paragraph{The source of each few-shot example consistently contributes more than its corresponding target.}
For each of the examined models, we notice that the source of each provided example is more influential than the corresponding target for generating the translation. This finding is consistent across language pairs. Aligning with findings in classical encoder-decoder MT models~\citep{ferrando-etal-2022-towards_opening_the_black_box-alti+, guerreiro_2023_hallucinations_in_llms}, where it was found that models tend to have higher source text contribution when translating into English than out of English, we find that the source contribution, both at the example and test source level, is higher for German to English than in English to German.

\paragraph{Training on parallel data reduces the impact of the provided examples on the translated sequence.}
We observe that the contributions of few-shot examples, particularly the first examples, are much greater for \textsc{Llama-2} than for both \textsc{Tower} models. One hypothesis is that the continued pretraining with parallel data on \textsc{Tower} makes it rely less on the examples since it is not required to ``learn'' the task ``on-the-fly''.
This leads to an interesting question: \textit{what if we replace the parallel data and instead only use monolingual data for multiple languages?} To investigate this, we examine the \textsc{Tower-Mono} model.\footnote{\textsc{Tower-Mono} was trained following the same training procedure as \textsc{Tower}~\citep{alves-2024-tower-llm}. The only difference to the former is that, instead of using 20B tokens of text split in 2/3 monolingual data and 1/3 parallel data, it was trained with 20B tokens of monolingual data.} Interestingly, we find that \textsc{Tower-Mono} behaves much more similarly to \textsc{Llama-2} than \textsc{Tower}. This suggests that continual pretraining with task-specific data may lead the model to rely less on examples to perform the task. Exploring how to train dedicated models to be better guided by in-context examples is an interesting direction for future work.

\paragraph{Close inspection of context contributions can uncover anomalous translations.} Previous works in neural MT have connected trends in context contributions, particularly low source contributions, to pathological translations such as hallucinations~\citep{ferrando-etal-2022-towards_opening_the_black_box-alti+,dale-etal-2023-halomi_hallucination_detection_MT, guerreiro_2023_hallucinations_in_llms}. Through close inspection of our analyzed samples, we indeed find a series of pathological translations. Figure~\ref{fig:anomalous_mt} presents one such example---here, the source contribution is particularly low, representing only about 25\% of the contribution of the first example; interestingly, the generated translation is, in fact, an exact copy of the translation from that first example. We provide additional examples in Appendix~\ref{app:anomalous_examples_high_level}. We will return to these and other salient cases in Section~\ref{sec:generation_stages} to examine how contributions evolve for such cases during the generation process.

\paragraph{A clear positional trend emerges in few-shot example contributions.}
Figure~\ref{fig:pos_bias_dataset_level_de_en_rel} shows a remarkable ``stair-like'' trend in the contribution of few-shot examples to the translated sequence. On average, the influence of each example appears to be strongly correlated with its position in the context, with earlier examples exhibiting higher contributions than later ones. This suggests there may be a positional bias in how the models leverage the provided examples during the translation process.

\section{Examining Positional Bias over the Provided Few-shot Examples}\label{sec:positional_bias}

Motivated by the findings from the previous section, we now closely inspect properties of the positional bias in few-shot example contributions.

\subsection{Are examples that occur early in the context more influential than later ones?}\label{subsec:positional_bias_analysis}
Here we perform a sample-level analysis to obtain a better understanding of the relationship between examples' contributions and their respective position. Specifically, we aim to explore whether there is a systematic and monotonic relationship between the order of few-shot examples and their contributions.

\begin{figure}[t]
    \centering
    \begin{subfigure}[b]{0.8\linewidth}
        \centering
        \includegraphics[width=\linewidth]{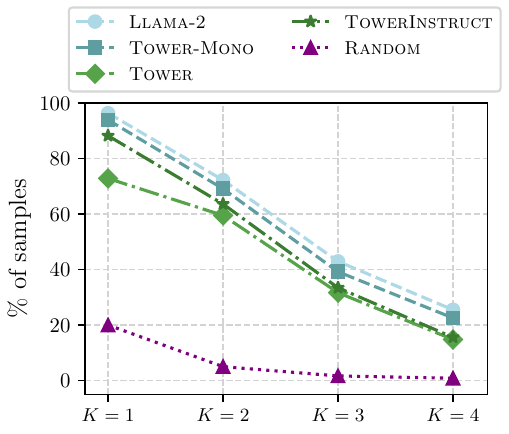}
        \caption{}
        \label{fig:pos_bias_analysis_de_en_rel}
    \end{subfigure}
    \begin{subfigure}[b]{0.8\linewidth}
        \centering
        \includegraphics[width=\linewidth]{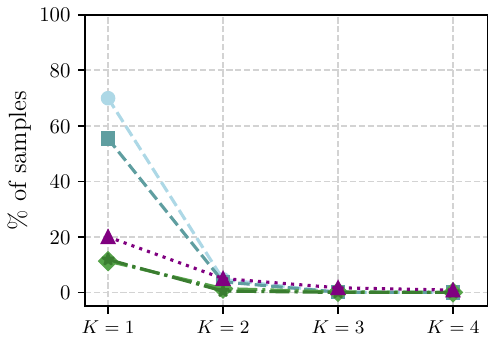}
        \caption{}
        \label{fig:pos_bias_break_de_en_relevant}
    \end{subfigure}
    \caption{Proportion of \texttt{de-en} samples that follow positional bias, for different values of $K$, in the (a)~original and (b)~\texttt{replace-last-ex} settings.}
    \label{fig:pos_bias_analysis_and_break_de_en}
\end{figure}

\subsubsection{Analysis setup} 

We examine whether the contributions of the first $K$ few-shot examples monotonically dominate the remaining $N-K$ examples, where $N$ is the total number of examples used in the context. In other words, for each sample, we check if the contributions of the first $K$ examples are sorted in descending order and if they are strictly higher than the contributions of the remaining $N-K$ examples.\footnote{We do not require the contributions of the remaining $N-K$ examples to be monotonically sorted.}
We consider different values of $K$ to represent different types of positional bias. For instance, when $K=1$, the first few-shot example attains the highest level of contribution. When $K=4$, the few-shot examples exhibit globally monotonic contributions, indicating a strong positional bias across all examples. Examples for each bias type are provided in Appendix~\ref{app:pos_bias_analysis}.

To quantify the prevalence of each type of positional bias, we measure the proportion of samples that satisfy the aforementioned condition for each value of $K$. We then compare these proportions to the probability, under a  permutation of the examples drawn uniformly at random (denoted as \textsc{Random}), of the first $K$ few-shot examples monotonically dominating the remaining $N-K$ examples, which is given as $p = (N-K)! / N!$.

\subsubsection{Results} 

We show results for German to English translation in Figure \ref{fig:pos_bias_analysis_de_en_rel}.\footnote{We include results for the rest language pairs examined in Appendix \ref{app:pos_bias_analysis}---trends are largely similar.}

\paragraph{Positional bias is prevalent and follows a monotonic pattern.}
Our analysis reveals that positional bias is significantly more common than the \textsc{Random} baseline for all values of $K$, suggesting that it is a prevalent phenomenon in the examined models. Additionally, we observe a monotonic relationship: the bias is more frequent for the first few examples than for later ones. This implies that the influence of positional bias gradually decreases as we move further down the context.

\paragraph{The bias is particularly stark for the first few-shot examples.}
All models tend to assign higher contribution to the first example, with this bias being more prevalent for models not trained on parallel data. For these models, over 95\% of the analyzed samples exhibit the highest contribution for the first example.\footnote{We remark again that the examples in the context are shuffled and not sorted by relevance to the source.} Models trained with parallel data, either through continued pretraining or additional finetuning, show a slight decrease in the first-example bias, but it remains significant compared to the \textsc{Random} baseline. 

The observed positional bias raises an important question: \textit{are contributions merely a function of position or are they connected to content of the context parts?} We will conduct two additional experiments in the next section to inspect this phenomenon closer.

\subsection{How strong is the positional bias?}
\label{subsec:breaking_pos_bias}
We now turn to a more detailed investigation of the positional trend we found in the results above. Specifically, we investigate how the introduction of other context parts and the relevance of the examples interact with the trend.

\begin{figure}[t]
    \centering
    \includegraphics[width=\columnwidth]{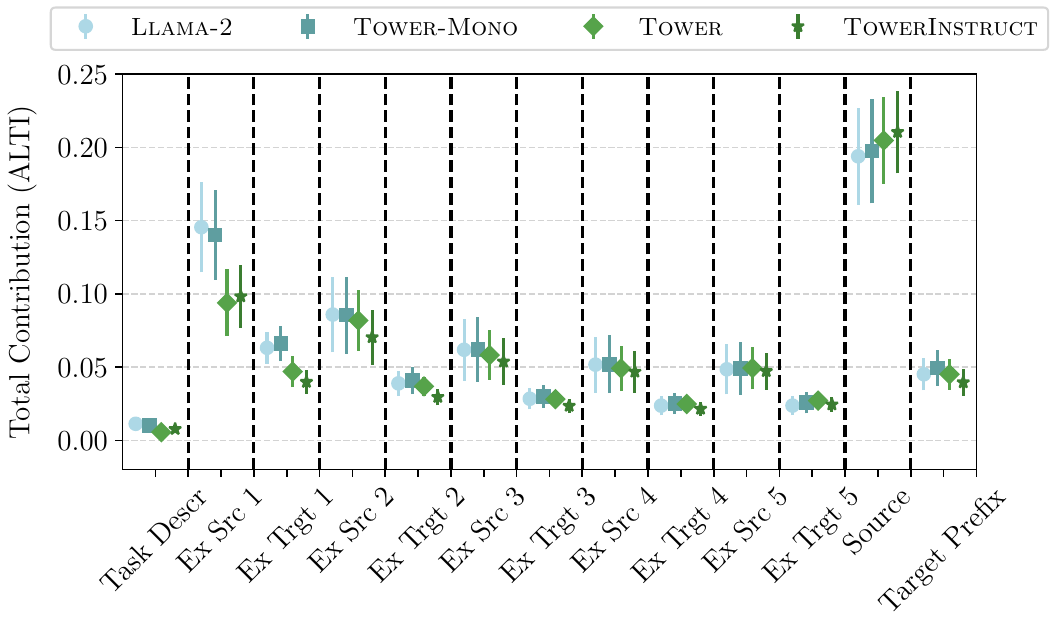}
    \captionsetup{width=\columnwidth}
    \caption{Illustration of context's part-level contributions, when the task description is added. Translation direction: \textit{German to English}}
\label{fig:preamble}
\end{figure}

\subsubsection{Is it all about position?} 

First, we examine the impact of adding a task description before the examples.\footnote{We can assume the "task description" as an additional part of the context. We use the following description template: \textit{Translate the following text from \texttt{[SRC\_LANG]} to \texttt{[TGT\_LANG]}\textbackslash n}.} If the bias is solely position-dependent, we might expect the task description to receive higher contribution due to its placement at the beginning of the context. This analysis will help us understand whether the positional bias is influenced by the nature of the content or if it is strictly position-based.

\paragraph{Task description receives minimal contribution despite its position.} The results of our first experiment, shown in Figure \ref{fig:preamble}, reveal that, despite appearing at the beginning of the input text, the task description receives significantly lower contribution compared to the examples and other parts of the context. This suggests that the positional bias is not merely a function of absolute position, but may rather depend on the nature of the content. Interestingly, even though a new part of context was added, the positional bias over the examples---``stair-like'' trend in the contributions---is still present.

\begin{figure*}[t]
    \centering
    \includegraphics[width=\textwidth]{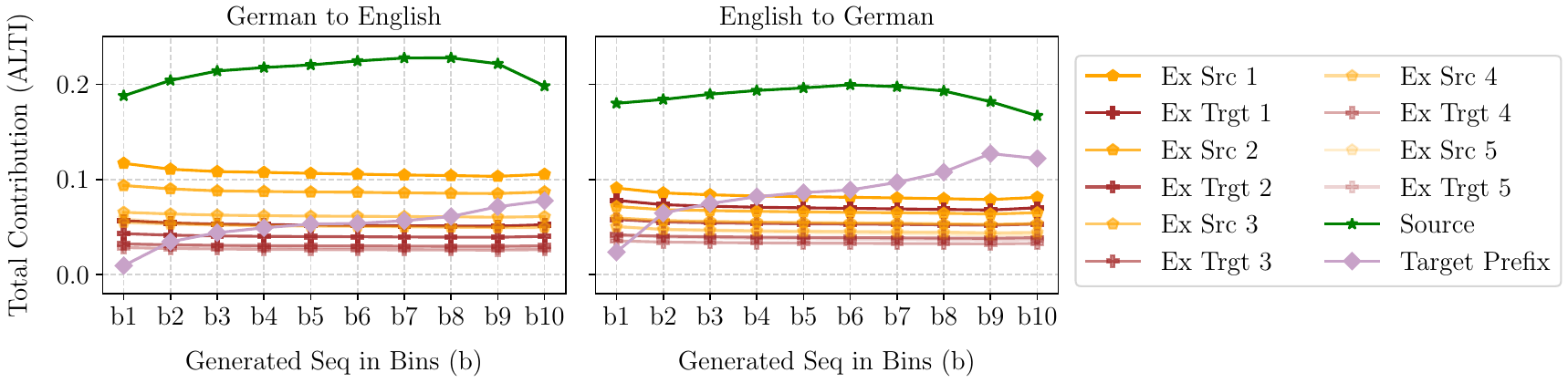}
    \captionsetup{width=\textwidth}
    \caption{Illustration of how context contributions evolve across different generation stages for the \textsc{Tower} model. Each generated bin accounts for 10\% of the generated sequence.}
\label{fig:dynamics_tower}
\end{figure*}

\subsubsection{Can relevance to the test example break the bias?}

We now investigate whether an overwhelmingly relevant example can break the positional bias, even when it appears later in the context.

To test this, we create an artificial setup---\texttt{replace-last-ex}---where a copy of the test example (source and translation) is placed as the last example in the context. Intuitively, if the model is shown a source text  along with its corresponding translation in the context, the most straightforward approach would be to copy the translation. As such, we expect the model to assign higher contribution to this last example, overriding the positional bias.

\paragraph{The bias is shrunk significantly.} Figure~\ref{fig:pos_bias_break_de_en_relevant} shows that this intervention significantly reduces the positional bias, particularly for the \textsc{Tower} and \textsc{TowerInstruct} models. In contrast, for models not trained on parallel data, the first example still contributes more than all other examples---even when a copy is present in the context---way more frequently than random chance. Interestingly, the bias is almost entirely broken for all other example positions. These findings suggest that while relevant content can indeed shrink the bias, the first examples influence the translation generation beyond simply ``solving the task.'' They likely provide additional cues, such as the language pair and expected output format, that shape the model's behavior.\vspace{6pt}

\section{How Do Context Contributions Evolve during the Generation Process?}
\label{sec:generation_stages}
In the previous sections, we examined which parts of the provided context have the greatest influence on the translated sequence. We now shift our focus to explore how these context contributions evolve across different stages of the generation process.

\subsection{Analysis setup}
To investigate this, we divide the generated sequence into 10 bins of equal length and compute the total contribution of each context part to each bin. We then average these contributions across samples to obtain a comprehensive view of how the influence of different context parts changes as the translation progresses.

\paragraph{Results.}
In Figure \ref{fig:dynamics_tower}, we present the average total contribution of each individual part to each generated bin, for the \textsc{Tower} models.

\paragraph{Relative ranking of context parts' contributions remains stable throughout generation.}
We observe that the relative ranking of contributions from different context parts is largely preserved throughout the generation process. Specifically, the source text consistently exhibits the highest contribution across all bins, followed by the few-shot examples in descending order of their position---this reinforces the notion of positional bias. The only exception to this pattern is the target prefix, which attains higher contribution as it grows in length. This is expected:~with a longer prefix, the model increasingly relies on the previously generated tokens to inform its predictions. Moreover, we also find a decrease in the source contribution at the last stage of generation, suggesting that the model relies less on the source when generating the final tokens. Interestingly, both these observations align with findings in traditional neural MT models, which have shown similar patterns in the relative contributions of source and target information during the generation process~\citep{voita-etal-2021-analyzing_source_target_contributions}.

\begin{table*}[t]
  \centering
  \footnotesize
  \begin{tabular}{lp{0.5\textwidth}l} 
    \cmidrule[\heavyrulewidth]{1-2}
      \centering
        \hlexampleone{\texttt{\textbf{E1}|SRC}} & Es gibt auch zwei schöne Parks in der Nähe, den Espanya Industrial Park und den Parc de Joan Miró. &\,\, \multirow{13}{*}[2mm]{\includegraphics[width=6cm]{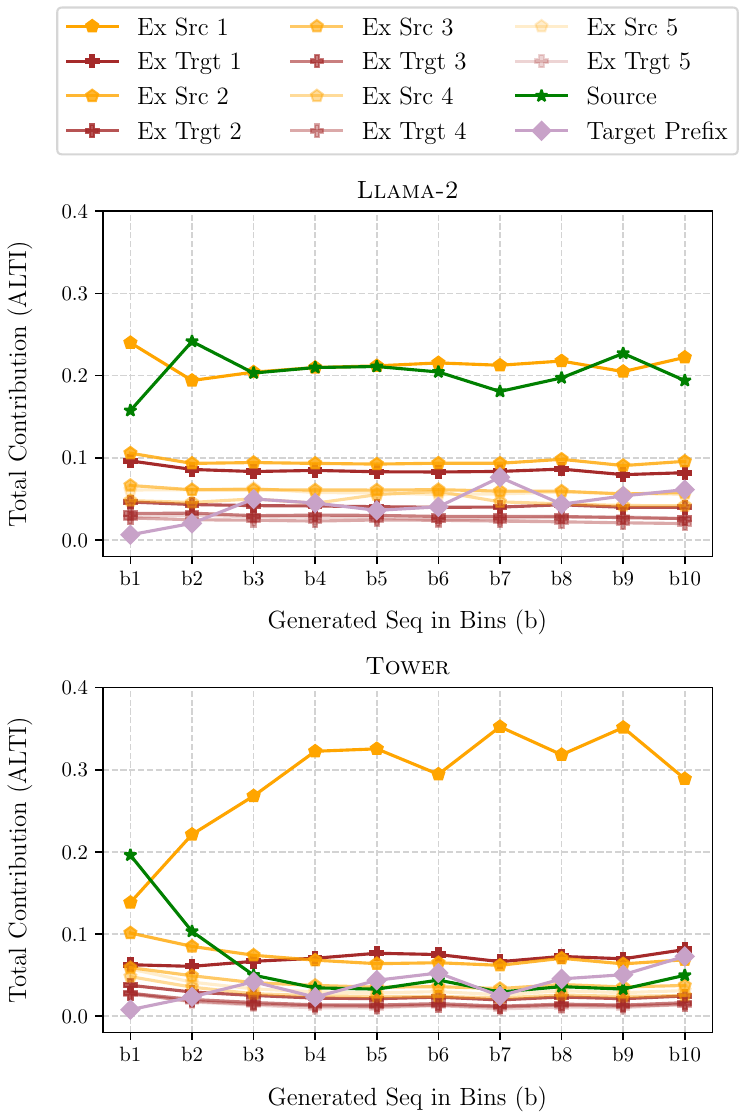}}\\
        \hlexampleonetgt{\texttt{\textbf{E1}|TGT}} & There are also two beautiful parks nearby, the Espanya Industrial Park and the Parc de Joan Miró.& \\\cdashlinelr{1-2}
        \hlexampleone{\texttt{\textbf{E2}|SRC}} & Das Frühstück ist im Preis (10 €) enthalten, es ist aber optional.& \\
        \hlexampleonetgt{\texttt{\textbf{E2}|TGT}} & Breakfast is included in the price (10 €), but it is optional.& \\\cdashlinelr{1-2}
        \hlexampleone{\texttt{\textbf{E3}|SRC}} &
        Es gibt auch kostenlose Internet 24/7 and WiFi in allen Zimmern.\\
        \hlexampleonetgt{\texttt{\textbf{E3}|TGT}} &
        There is also free internet 24/7and wifi in all rooms. & \\\cdashlinelr{1-2}
        \hlexampleone{\texttt{\textbf{E4}|SRC}} &
        Bisher gibt es noch keine Bewertungen für S-Plus Company!& \\
        \hlexampleonetgt{\texttt{\textbf{E4}|TGT}} & 
        There are no reviews for S-Plus Company yet! & \\\cdashlinelr{1-2}
        \hlexampleone{\texttt{\textbf{E5}|SRC}} & 
        Die Größe der Wohnung ist 15 m2, es ist klein, aber sehr gemütlich. & \\
        \hlexampleonetgt{\texttt{\textbf{E5}|TGT}} &
        The size of the apartment is 15 m2, it's small but very cosy.& \\\cdashlinelr{1-2}
        \hlsrc{\texttt{\textbf{SRC}}} & Die gibt es zwar auch (anscheinend?) bei den MarathonPlus Reifen, aber der Großteil ist schon breiter.& \\\cdashlinelr{1-2}
        \multicolumn{2}{l}{\textcolor{black!70}{\textit{\textsc{Llama-2}}} \textcolor{darkgreen}{\ding{51}}}\\ 
        \hlmt{\texttt{\textbf{MT}}} & There are also (apparently?) at Marathon Plus Tyres, but the majority is wider.\\\cdashlinelr{1-2}
        \multicolumn{2}{l}{\textcolor{black!70}{\textit{\textsc{Tower}}} \textcolor{red}{\ding{55}}}\\ 
        \hlmt{\texttt{\textbf{MT}}} & There are also two beautiful parks nearby, the Espanya Industrial Park and the Parc de Joan Miró.\\\cmidrule[\heavyrulewidth]{1-2}

  \end{tabular}
  \caption{Illustration of an example exhibiting anomalous source contributions for \textsc{Tower} --- which hallucinates,  followed by \textsc{Llama-2}'s contributions, which performs normally.}
  \label{tab:anomalous_example_985}
\end{table*}

\paragraph{Translation direction impacts the evolution of context contributions.}
While the overall ranking of context part contributions remains similar, we observe notable differences when translating into or out of English. As noted earlier in Section~\ref{sec1:high_level_analysis_segment_contributions}, the source contribution is higher when translating into English~(\texttt{de-en}) compared to when translating out of English~(\texttt{en-de}). Interestingly, in \texttt{de-en} translation, the source of each example also consistently contributes more than its corresponding target, resulting in a ``stacked'' appearance of source contributions---the contribution from any example's source is bigger than that of any example's target text. In contrast, \texttt{en-de} translation exhibits an alternating contribution ranking, with the source and target of each example interleaved (e.g., \texttt{src} example 1 > \texttt{tgt} example 1 > \texttt{src} example 2 > \texttt{tgt} example 2, and so on). Moreover, we also observe that the target prefix contribution grows much more steeply in \texttt{en-de} than in \texttt{de-en}, suggesting that when translating a non-English text, the model relies more heavily on the context (examples and source) throughout the generation process.

\paragraph{Highlighting the importance of source-part contributions in anomalous cases.} Building on our findings from Section \ref{sec1:high_level_analysis_segment_contributions}, which showed that close inspection of context contributions can potentially uncover anomalous translations, we further analyze such cases in terms of how context contributions evolve during the generation process. We compare the behavior of \textsc{Llama-2} and \textsc{Tower} models using the example presented in Table~\ref{tab:anomalous_example_985}~(the same presented in Section~\ref{sec1:high_level_analysis_segment_contributions}). For \textsc{Llama-2}, which generates a correct translation, the context contribution trends align with the average case for German to English translation (see Figure \ref{fig:dynamics_remaining} in Appendix \ref{app:generation_stages_plots}). In contrast, \textsc{Tower}, which produces an incorrect translation by copying the first example, exhibits anomalous contribution trends (compared to Figure~\ref{fig:dynamics_tower}). Specifically, we observe a steeply increasing contribution from the first example, while the source contribution decreases significantly, highlighting the copying behavior. Additional salient cases are discussed in Appendix~\ref{app:generation_stages_anamalous_examples}.\footnote{Here, we not only provide examples of other hallucinations, but also of other correct translations for which the context contributions follow interesting non-typical patterns.} Crucially, we find that in such cases, \textit{source contributions}---both at the example and test source levels---can \textit{potentially} indicate \textit{pathological translations} and also provide insights into the factors driving the generation. 
\begin{table}[h]
\footnotesize
\centering
    \begin{tabular}{ccc}
        \toprule
        \textbf{Language Pair} & \textbf{Model}  &\textbf{AUROC}\\
        \midrule
        \texttt{en-ru} & \textsc{llama-2}&$52.3$\\
        \texttt{de-en} & \textsc{Tower}&$97.3$\\
        \texttt{en-ru} &  \textsc{Tower} &$88.7$\\
        \hline
    \end{tabular}
    \caption{AUROC of low source contribution scores.}
    \label{tab:auroc}
\end{table}
\paragraph{Low source contributions are, \textit{in some cases}, predictive of hallucinations.} Our previous observations may potentially align well with previous neural MT research linking pathological translations to low source contributions~\citep{ferrando-etal-2022-towards_opening_the_black_box-alti+,dale-etal-2023-halomi_hallucination_detection_MT, guerreiro_2023_hallucinations_in_llms}. Note again that classical encoder-decoder MT models and large language models (LLMs) are distinct in terms of the parts of context they often support: in classical encoder-decoder NMT models, the "context" for generation typically comprises only the source sentence and previously generated tokens; LLMs, however, often maintain a much broader context, potentially including various other relevant information. This distinction means that low source contribution in LLMs may not be so predictive of pathological translations, as the model might be drawing from other relevant contextual information. To explore this further, we conduct a quantitative analysis to assess the extent to which low-source contribution can be associated with hallucinations. Initially, for each model and language pair combination, we identify instances of "fully-detached" hallucinations by annotating the generated translations using the \textsc{Llama-3-70B-Instruct} model~\cite{dubey2024llama3herdmodels}, following the exact approach outlined by~\citealp{benkirane2024machinetranslationhallucinationdetection}.\footnote{In this paper, the authors show that this LLM can achieve performance comparable or even better than previously proposed detectors.} For each model-language pair combination for which we observed a reasonable number\footnote{We provide further quantitative resuls on the number of detected hallucinations in Appendix~\ref{app:quantitative}.} of "fully-detached" hallucinations, we report the AUROC of the low source contribution score in Table~\ref{tab:auroc}. Our findings, suggest that while for \textsc{Tower} low source contributions are particularly associated with hallucinations, it is not the case for \textsc{llama2}. Upon closer inspection, we find that the low source contribution is particularly predictive of hallucinations that come in the form of exact copies of the provided few-shot examples\footnote{\textsc{Tower'}s pathological translations are usually copies of the few-shot examples, while this is not the case for \textsc{Llama2}.}. Investigating these trends further, not only in machine translation but also in other tasks where context is relevant, is an interesting direction for future research.

\section{Conclusion}\label{sec:conclusion} 
 
We have comprehensively studied context contributions in LLM-based MT using the general purpose \textsc{Llama-2} and translation-specialized \textsc{Tower} models, exploring a broad range of key aspects, including investigating how different parts of context contribute to generated translations, and how these contributions evolve during the generation process.

Our findings reveal a strong positional bias, where earlier few-shot examples in the context have higher contributions to the translated sequence, both at the sentence level and across different generation stages. Interestingly, our experiments show that this bias is shrunk by continuous pretraining on task-specific data. Moreover, we reveal that the source part of each few-shot example has higher contribution compared to its corresponding target, irrespective of the translation direction. Finally, we stress the importance of source-part contributions by demonstrating that anomalous contributions can potentially uncover pathological translations, such as hallucinations. We believe our work not only provides insights into the internal workings of LLM-based MT, but also draws important connections to classical encoder-decoder NMT models. 

To support future research on this topic, we are
open-sourcing our code and releasing all data used in our analysis.

\section*{Limitations}
While our study provides a valuable insight of how context is utilized by LLMs in MT, there are a few limitations that should be acknowledged. 

Firstly, due to limitations in terms of computational resources paired with the fact that the ALTI method employed in our study can be computationally intensive, we restricted our analysis to 7B parameter models. This constraint raises the question of whether our findings still hold true when larger LLMs are considered, making it a potential direction for future studying.

Secondly, it should be noted that we  focused exclusively on \textsc{Llama}-based models, particularly aiming on analyzing the \textsc{Tower}-family of models, which are specifically oriented for MT. This selection enabled us to study how continued pretraining and finetuning on task-specific data impacts context utilization. However, this decision makes it so that it is still unclear whether our findings generalize to other LLM families.

Despite these limitations, we believe our study can lead to a better understanding of the dynamics of context utilization in LLM-based MT, providing key insights that can motivate future work on the field and inspire other research directions. 

\section*{Ethical Considerations \& Potential Risks}
Utilizing LLMs for MT might raise  potential risks that should be pointed out, particularly regarding pathological translations and the ethical usage of contextual data.

Firstly, one of the critical risks which arises when using LLMs for MT is the phenomenon of pathological translations, such as hallucinations. As our study reveals, anomalous context contributions can potentially indicate these pathological translations, especially when low reliance on the source text is noticed. Despite the potential of detecting these pathological translations, their occurrence remains an important concern, as misinterpretations and incorrect translations might lead to significant consequences in specific domains such as healthcare, law etc. Thus ensuring that LLMs provide reliable translations is crucial.

Secondly, the reliance of LLMs in specific parts of the context when translating, introduces ethical considerations that should be taken into account regarding the choice of  some context parts, such as the few-shot examples. The provided context might contain biases and misleading or inappropriate content and as a result this might be propagated into the generated translations. 
Our research can significantly contribute to mitigate this risk by identifying which parts of the provided context are responsible for propagating biases or inappropriate content to the translated sequence. 

To conclude, addressing these risks and ethical considerations is important to foster a better usage of these systems and prevent  potential harms.

\section*{Acknowledgements}
This work was supported by the Portuguese Recovery and Resilience Plan through project C645008882-00000055 (Center for Responsible AI), by EU's Horizon Europe Research and Innovation Actions (UTTER, contract 101070631), by the project DECOLLAGE (ERC-2022-CoG 101088763), and by Fundação para a Ciência e Tecnologia through contract UIDB/50008/2020.

\bibliography{custom}

\appendix

\section{Further Details on Experimental Setup}
\label{app:Experimental_setup_details}

\subsection{Few-shot setting \& Prompt selection}
\label{app:few_shot_selection}

We conduct our experiments using the few-shot examples provided by \citealt{henry_gpt-mt-2023}, which were selected to be of high-quality and relevant to the source.

Following prior work \citep{zhang-icml-2023-prompting-llms-for-mt}, we use the in-context template illustrated in Table~\ref{tab:prompt_templates}.

\begin{table}[h]
    \centering
    \begin{tabular}{ll}
    \texttt{SRC\_LANG}:  & \hlexampleone{\texttt{\textbf{E1}|SRC}} \\
    \texttt{TGT\_LANG}:  & \hlexampleonetgt{\texttt{\textbf{E1}|TGT}} \\
    \texttt{SRC\_LANG}:  & \hlexampleone{\texttt{\textbf{E2}|SRC}} \\
    \texttt{TGT\_LANG}:  & \hlexampleonetgt{\texttt{\textbf{E2}|TGT}} \\
    \texttt{[...]}                & \\
    \texttt{SRC\_LANG}:  & \hlsrc{\texttt{\textbf{SRC}}} \\
    \texttt{TGT\_LANG}:  & 
    \end{tabular}
    \caption{Prompt template for few-shot inference.}
    \label{tab:prompt_templates}
\end{table}

\subsection{Filtering details}
\label{app:filtering_details}

Due to our resource constraints, coupled with the high GPU memory requirements of the attribution method when applied to a 7B parameter model, we had to filter samples with large context length. More specifically, we exclude samples exceeding 400 tokens, when considering the concatenation of the input prompt with the generated sequence. We additionally filter out the samples for which the generated sequence does not exceed the length of 10 tokens.\footnote{In our analysis in Section \ref{sec:generation_stages}, we separate the generated sequences into 10 bins.} We report the sizes of the sets---over 1000 samples for each language pair---examined in our analysis in Table \ref{tab:sample_sizes}.

\begin{table}[h]
\centering
    \begin{tabular}{cc}
        \toprule
        \textbf{Language Pair} & \textbf{Sample Size}\\ 
        \midrule
        De-En & 1021 \\
        Ru-En & 1017 \\
        En-De & 1174\\
        En-Ru & 1107\\
        \hline
    \end{tabular}
    
    \caption{Sample sizes for each language pair considered in our analysis.}
    \label{tab:sample_sizes}
\end{table}

\subsection{Evaluation Details}
\label{app:evaluation_details}

We evaluate the models used in our work on all language directions examined to ensure high translation quality. We report BLEU \cite{papineni-2002-bleu}, COMET-22 \cite{rei-2022-comet-22}, and COMETKiwi \cite{rei-acl-2022-comet-kiwi} in Table 
\ref{tab:eval_metrics}.

\subsection{Inference} 
We used greedy decoding at inference time, setting 300 tokens as the maximum length for the generated sequence.

\subsection{Hardware specifications}
All our experiments were conducted using 3 NVIDIA RTX A6000 GPUs.

\subsection{Discussion on artifacts}
The data used for analysis in this paper was initially released for the WMT22 General MT task~\citep{kocmi-etal-2022-findings} and can be freely used for research purposes. All translation demonstrations (few-shot examples) used in our paper were released in~\cite{henry_gpt-mt-2023} under a MIT license. 

Our code was developed on top of original ALTI repositories~\citep{ferrando-etal-2022-towards_opening_the_black_box-alti+, ferrando-etal-2023-explaining}, which have been released under Apache-2.0 License.

\begin{table*}[t]
    \centering
    \begin{tabular}{lcccccccccccccccc}
        \toprule
        \multicolumn{1}{c}{} & 
        \multicolumn{3}{c}{De-En } & \multicolumn{3}{c}{En-De} 
        \\
        \cmidrule(lr){2-4}
        \cmidrule(lr){5-8}
          & BLEU & COMET-22 & COMETKiwi & BLEU & COMET-22 & COMETKiwi \\
        \midrule
          \textsc{Llama-2}&$28.42$& $82.25$&$78.82$&       $21.12$& $78.79$&$74.95$\\
          \textsc{Tower-Mono}&$28.19$& $82.45$ &$78.90$&    $23.42$& $80.99$& $77.88$\\
          \textsc{Tower}& $30.19$&$83.22$&$79.60$
          & $29.39$& $84.40$& $81.58$\\
          \textsc{TowerInstruct}&$35.24$&$85.72$&$81.43$
          & $42.66$& $88.11$& $83.11$\\
        \bottomrule
       \multicolumn{1}{c}{} & 
        \multicolumn{3}{c}{Ru-En } & \multicolumn{3}{c}{En-Ru} 
        \\
        \cmidrule(lr){2-4}
        \cmidrule(lr){5-8}
          & BLEU & COMET-22 & COMETKiwi & BLEU & COMET-22 & COMETKiwi \\
        \midrule
          \textsc{Llama-2}&$32.99$& $82.53$&$78.84$&       $20.03$& $80.78$&$76.80$\\
          \textsc{Tower-Mono}&$33.47$& $83.04$ &$79.16$&    $23.19$& $83.26$& $79.31$\\
          \textsc{Tower}& $37.78$&$83.84$&$79.79$
          & $28.33$& $86.10$& $82.03$\\
          \textsc{TowerInstruct}&$44.48$&$86.53$&$81.51$
          & $40.02$& $89.72$& $83.41$\\
        \bottomrule
    \end{tabular}
        \caption{Translation performance of each examined model on the WMT22 test set.}
    \label{tab:eval_metrics}
\end{table*}

\begin{figure*}[ht]
    \centering
    \includegraphics[width=\textwidth]{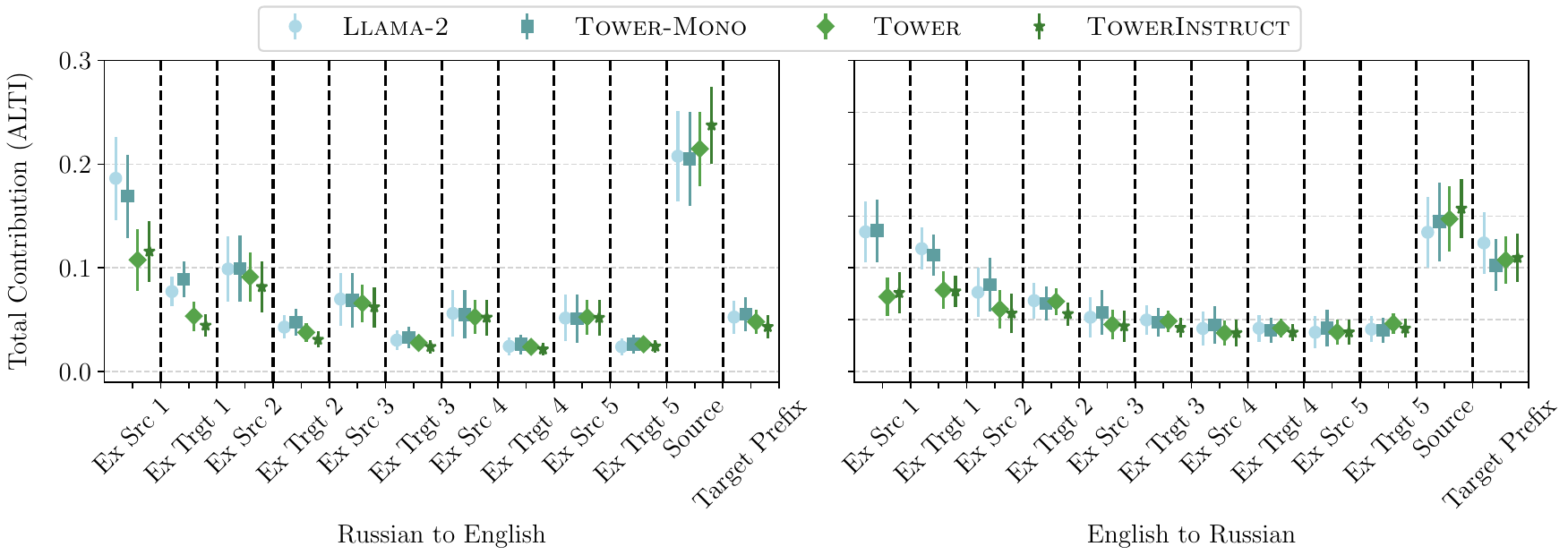}
    \captionsetup{width=\textwidth}
    \caption{Illustration of context's part-level contributions to the translated sequence, for all the examined models.}
\label{fig:pos_bias_dataset_level_ru_en_rel}
\end{figure*}

\begin{figure*}[ht]
    \centering
    \includegraphics[width=\textwidth]{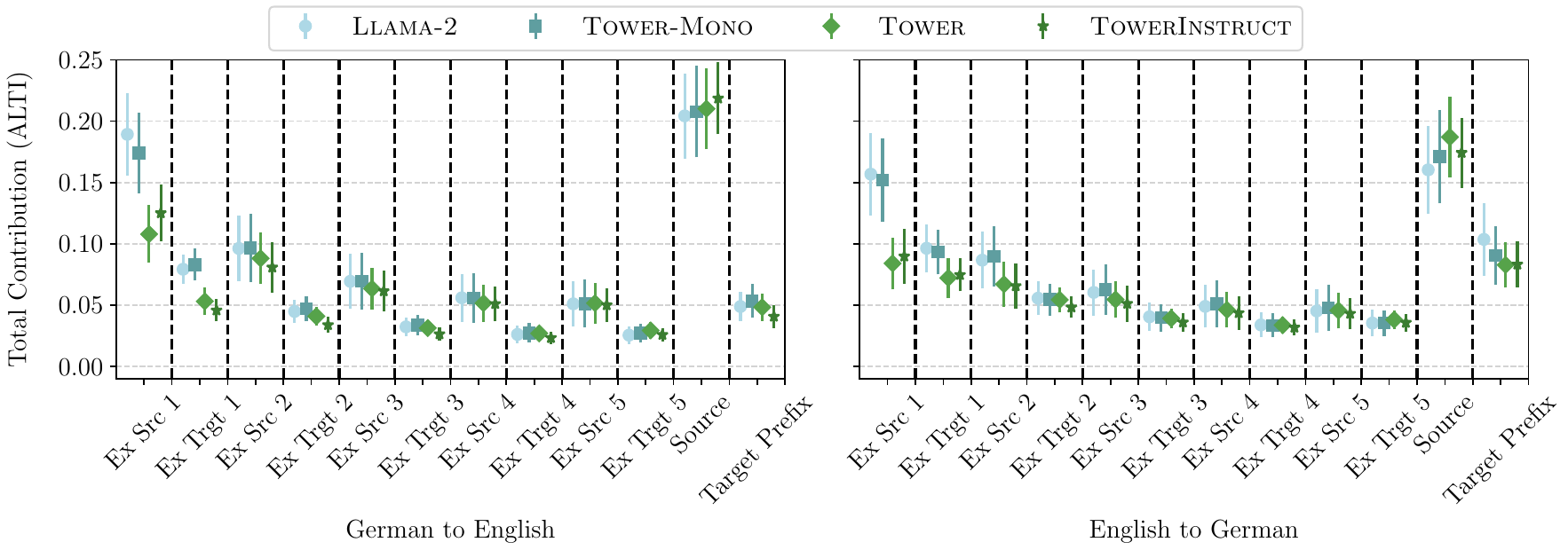}
    \captionsetup{width=\textwidth}
    \caption{Illustration of context's part-level contributions to the translated sequence, when reshuffling the order of provided few-shot examples.} \label{fig:high_level_contr_reshuffle}
\end{figure*}

\begin{figure*}[ht]
    \centering
    \includegraphics[width=\textwidth]{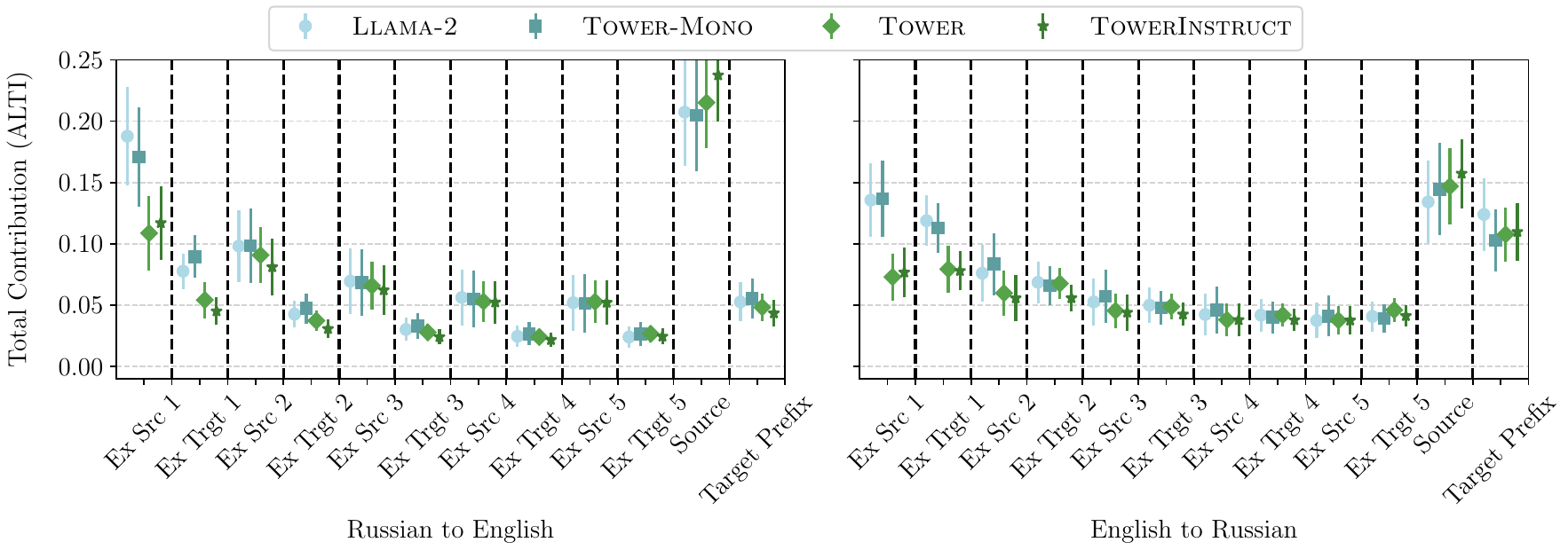}
    \captionsetup{width=\textwidth}
    \caption{Illustration of context's part-level contributions to the translated sequence, when reshuffling the order of provided few-shot examples.} \label{fig:high_level_contr_reshuffle_russian}
\end{figure*}

\begin{figure*}[t]
    \centering
\begin{tikzpicture}[scale=0.5, every node/.style={font=\footnotesize, inner sep=0,outer sep=0}]
\node [] (A) at (-0.75,5.5) {\hlexampleone{\texttt{\textbf{E1}|SRC}} Leider konnten wir keine Shops finden,die Folgendes anbieten: Buch mit ISBN '9789635487899'.};
\node [below=0.5cm of A.west,anchor=west] (B) {\hlexampleonetgt{\texttt{\textbf{E1}|TGT}} \ul{Unfortunately we could not find any stores offering the Book with ISBN '9789635487899'.}};
\node [below=0.5cm of B.west,anchor=west] (C) {\hlexampleone{\texttt{\textbf{E2}|SRC}} Deezer auf Xbox One – Deezer Support};
\node [below=0.5cm of C.west,anchor=west] (D) {\hlexampleonetgt{\texttt{\textbf{E2}|TGT}} Deezer on Xbox One – Deezer Support};
\node [below=0.5cm of D.west,anchor=west] (E) {\hlexampleone{\texttt{\textbf{E3}|SRC}} Installieren Sie die Mercedes PRO Adapter App2 auf Ihrem Smartphone.};
\node [below=0.5cm of E.west,anchor=west] (F) {\hlexampleonetgt{\texttt{\textbf{E3}|TGT}} Install the Mercedes PRO Adapter App2 on your smartphone.};

\node [below=0.5cm of F.west,anchor=west] (G) {\hlexampleone{\texttt{\textbf{E4}|SRC}} Spielen MetalStorm: Online auf Ihrem mobilen Gerät.};
\node [below=0.5cm of G.west,anchor=west] (H) {\hlexampleonetgt{\texttt{\textbf{E4}|TGT}} Play MetalStorm: Online on your mobile device.};

\node [below=0.5cm of H.west,anchor=west] (I) {\hlexampleone{\texttt{\textbf{E5}|SRC}} support@vivago.com (Technischer Support)};
\node [below=0.5cm of I.west,anchor=west] (J) {\hlexampleonetgt{\texttt{\textbf{E5}|TGT}} support@vivago.com (Technical Support)};

\node [below=0.5cm of J.west,anchor=west] (L) {\hspace{14pt}\hlsrc{\texttt{\textbf{SRC}}} Leider warte ich vergeblich auf die email von  ihrem Support.};
\node [below=0.5cm of L.west,anchor=west] (M) {\hspace{16pt} \hlmt{\texttt{\textbf{MT}}} \ul{Unfortunately, we could not find any stores offering the Book with ISBN '9789635487899'.}};

\draw (15,7.25) node {\textsf{Contribution Ratio}};
\draw (15,6.5) node {\textsf{to \hlexampleone{\texttt{\textbf{E1}|SRC}}}};
\draw[thick, dashed, gray!60] (14,-6) -- (14,6);
\draw[thick, dashed, gray!60] (14,6) -- (14,-6) node[anchor=north, black] {$0$};
\draw[thick, dashed, gray!60] (15,6) -- (15,-6) node[anchor=north, black] {$0.5$};
\draw[thick, dashed, gray!60] (16,6) -- (16,-6) node[anchor=north, black] {$1$};
\draw[thick, dashed, gray!60] (14,5.5) -- (16,5.5);
\filldraw[blue, draw=gray!20] (14,5.25) rectangle (16,5.75); 
\draw[thick, dashed, gray!60] (14,4.5) -- (16,4.5);
\filldraw[blue, draw=gray!20] (14,4.25) rectangle (14.916,4.75);
\draw[thick, dashed, gray!60] (14,3.5) -- (16,3.5);
\filldraw[blue, draw=gray!20] (14,3.25) rectangle (14.475,3.75);
\draw[thick, dashed, gray!60] (14,2.5) -- (16,2.5);
\filldraw[blue, draw=gray!20] (14,2.25) rectangle (14.158,2.75);
\draw[thick, dashed, gray!60] (14,1.5) -- (16,1.5);
\filldraw[blue, draw=gray!20] (14,1.25) rectangle (14.383,1.75);  
\draw[thick, dashed, gray!60] (14,0.5) -- (16,0.5);
\filldraw[blue, draw=gray!20] (14,0.25) rectangle (14.133,0.75);  
\draw[thick, dashed, gray!60] (14,-0.5) -- (16,-0.5);
\filldraw[blue, draw=gray!20] (14,-0.75) rectangle (14.241,-0.25);  
\draw[thick, dashed, gray!60] (14,-1.5) -- (16,-1.5);
\filldraw[blue, draw=gray!20] (14,-1.75) rectangle (14.1,-1.25);  
\draw[thick, dashed, gray!60] (14,-2.5) -- (16,-2.5);
\filldraw[blue, draw=gray!20] (14,-2.75) rectangle (14.225,-2.25);  
\draw[thick, dashed, gray!60] (14,-3.5) -- (16,-3.5);
\filldraw[blue, draw=gray!20] (14,-3.75) rectangle (14.1,-3.25);  
\draw[thick, dashed, gray!60] (14,-4.5) -- (16,-4.5);
\filldraw[blue, draw=gray!20] (14,-4.75) rectangle (14.433,-4.25);  
\draw[thick, dashed, gray!60] (14,-5.5) -- (16,-5.5);
\filldraw[blue, draw=gray!20] (14,-5.75) rectangle (14.758,-5.25);  

\end{tikzpicture}
    \caption{Example of anomalous source contributions for \textsc{Tower} which hallucinates, copying information from the first example. We show contribution ratios to \hlexampleone{\texttt{\textbf{E1}|SRC}}---$1$ being the contribution of \hlexampleone{\texttt{\textbf{E1}|SRC}}.}
    \label{fig:hal_t_ex_575}
\end{figure*}

\begin{figure*}[t]
    \centering
\begin{tikzpicture}[scale=0.5, every node/.style={font=\footnotesize, inner sep=0,outer sep=0}]
\node [] (A) at (-0.75,5.5) {\hlexampleone{\texttt{\textbf{E1}|SRC}} Wir wünschen Ihnen einen angenehmen Aufenthalt in Maribor.};
\node [below=0.5cm of A.west,anchor=west] (B) {\hlexampleonetgt{\texttt{\textbf{E1}|TGT}} We wish you a pleasant stay in Maribor.};
\node [below=0.5cm of B.west,anchor=west] (C){\hlexampleone{\texttt{\textbf{E2}|SRC}} Wir wünschen Ihnen einen angenehmen Aufenthalt in Olomouc.};
\node [below=0.5cm of C.west,anchor=west] (D) {\hlexampleonetgt{\texttt{\textbf{E2}|TGT}} We wish you a pleasant stay in Olomouc.};
\node [below=0.5cm of D.west,anchor=west] (E) {\hlexampleone{\texttt{\textbf{E3}|SRC}} Wir wünschen Ihnen einen angenehmen Aufenthalt in Debrecen.};
\node [below=0.5cm of E.west,anchor=west] (F) {\hlexampleonetgt{\texttt{\textbf{E3}|TGT}} We wish you a pleasant stay in Debrecen.};

\node [below=0.5cm of F.west,anchor=west] (G) {\hlexampleone{\texttt{\textbf{E4}|SRC}} Wir wünschen Ihnen einen angenehmen Aufenthalt in Poznan.};
\node [below=0.5cm of G.west,anchor=west] (H) {\hlexampleonetgt{\texttt{\textbf{E4}|TGT}} We wish you a pleasant stay in Poznan.};

\node [below=0.5cm of H.west,anchor=west] (I) {\hlexampleone{\texttt{\textbf{E5}|SRC}} Busbud hilft Ihnen, einen Bus von Lübeck nach Wismar zu finden.};
\node [below=0.5cm of I.west,anchor=west] (J) {\hlexampleonetgt{\texttt{\textbf{E5}|TGT}}\ul{ Busbud helps you find a bus from Lübeck to Wismar.}};

\node [below=0.5cm of J.west,anchor=west] (L) {\hspace{14pt}\hlsrc{\texttt{\textbf{SRC}}} Wir verraten Ihnen, wo Sie im Raum Lübeck doch noch einen Weihnachtsbraten herbekommen.};
\node [below=0.5cm of L.west,anchor=west] (M) {\hspace{16pt} \hlmt{\texttt{\textbf{MT}}} \ul{ Busbud helps you find a bus from Lübeck to Wismar.}};

\draw (19,7.25) node {\textsf{Contribution Ratio}};
\draw (19,6.5) node {\textsf{to \hlmt{\texttt{\textbf{MT}}}}};
\draw[thick, dashed, gray!60] (18,-6) -- (18,6);
\draw[thick, dashed, gray!60] (18,6) -- (18,-6) node[anchor=north, black] {$0$};
\draw[thick, dashed, gray!60] (19,6) -- (19,-6) node[anchor=north, black] {$0.5$};
\draw[thick, dashed, gray!60] (20,6) -- (20,-6) node[anchor=north, black] {$1$};
\draw[thick, dashed, gray!60] (18,5.5) -- (20,5.5);
\filldraw[blue, draw=gray!20] (18,5.25) rectangle (19.45,5.75); 
\draw[thick, dashed, gray!60] (18,4.5) -- (20,4.5);
\filldraw[blue, draw=gray!20] (18,4.25) rectangle (18.54,4.75);
\draw[thick, dashed, gray!60] (18,3.5) -- (20,3.5);
\filldraw[blue, draw=gray!20] (18,3.25) rectangle (18.89,3.75);
\draw[thick, dashed, gray!60] (18,2.5) -- (20,2.5);
\filldraw[blue, draw=gray!20] (18,2.25) rectangle (18.44,2.75);
\draw[thick, dashed, gray!60] (18,1.5) -- (20,1.5);
\filldraw[blue, draw=gray!20] (18,1.25) rectangle (18.62,1.75);  
\draw[thick, dashed, gray!60] (18,0.5) -- (20,0.5);
\filldraw[blue, draw=gray!20] (18,0.25) rectangle (18.34,0.75);  
\draw[thick, dashed, gray!60] (18,-0.5) -- (20,-0.5);
\filldraw[blue, draw=gray!20] (18,-0.75) rectangle (18.458,-0.25);  
\draw[thick, dashed, gray!60] (18,-1.5) -- (20,-1.5);
\filldraw[blue, draw=gray!20] (18,-1.75) rectangle (18.28,-1.25);  
\draw[thick, dashed, gray!60] (18,-2.5) -- (20,-2.5);
\filldraw[blue, draw=gray!20] (18,-2.75) rectangle (19.75,-2.25);  
\draw[thick, dashed, gray!60] (18,-3.5) -- (20,-3.5);
\filldraw[blue, draw=gray!20] (18,-3.75) rectangle (19.08,-3.25);  
\draw[thick, dashed, gray!60] (18,-4.5) -- (20,-4.5);
\filldraw[blue, draw=gray!20] (18,-4.75) rectangle (18.9,-4.25);  
\draw[thick, dashed, gray!60] (18,-5.5) -- (20,-5.5);
\filldraw[blue, draw=gray!20] (18,-5.75) rectangle (20,-5.25);  

\end{tikzpicture}
    \caption{Example of anomalous source contributions for \textsc{Tower} which hallucinates, copying information from the last example. We show contribution ratios to \hlmt{\texttt{\textbf{MT}}}---$1$ being the contribution of \hlmt{\texttt{\textbf{MT}}}.}
    \label{fig:hal_t_ex_653}
\end{figure*}

\begin{figure*}[t]
    \centering
    \begin{subfigure}[t]{0.49\textwidth}
      \includegraphics[width=\textwidth]{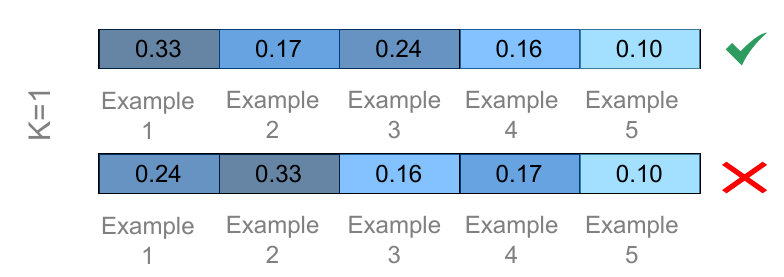}
        \caption{The top sample follows the examined positional bias ($K=1$) as the first example attains the highest contribution. The bottom sample does not follow the bias, as the second example has greater contribution than the first.}
    \label{fig:pos_type_ex_k1}
    \end{subfigure}
      \hfill
      \begin{subfigure}[t]{0.49\textwidth}
    \includegraphics[width=\textwidth]{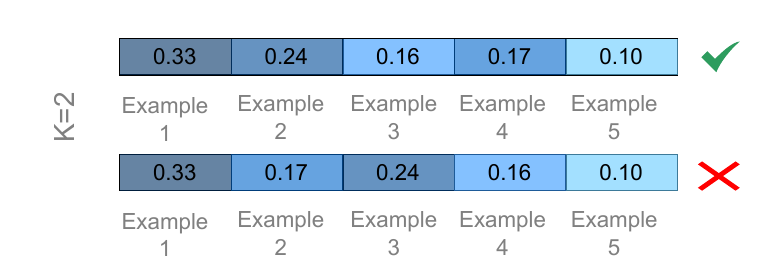}
            \caption{The top sample follows the examined positional bias ($K=2$) as the first two examples monotonically dominate the remaining three and the last three have lower contributions than the first two. Note that the last three examples do not necessarily exhibit sorted contributions  in decreasing order. The bottom sample does not follow the bias, as the third example has greater contribution than the second.}
        \label{fig:pos_type_ex_k2}
      \end{subfigure}

\vskip\baselineskip
    \begin{subfigure}[t]{0.49\textwidth}
      \includegraphics[width=\textwidth]{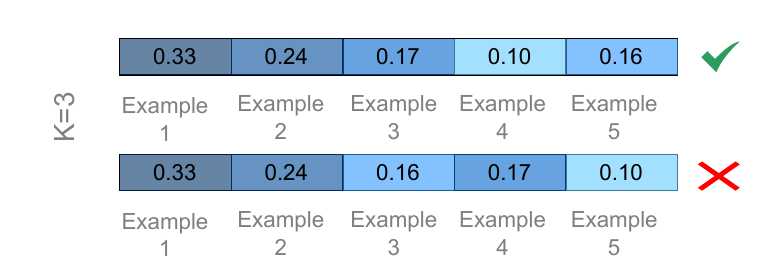}
        \caption{The top sample follows the examined positional bias ($K=3$) as the first three examples monotonically dominate the remaining two and the last two have lower contributions than the first three. Note that the last two examples do not necessarily exhibit sorted contributions in decreasing order. The bottom sample does not follow the bias, as the fourth example has greater contribution than the third.}
    \label{fig:pos_type_ex_k3}
    \end{subfigure}
  \hfill
  \begin{subfigure}[t]{0.49\textwidth}
    \includegraphics[width=\textwidth]{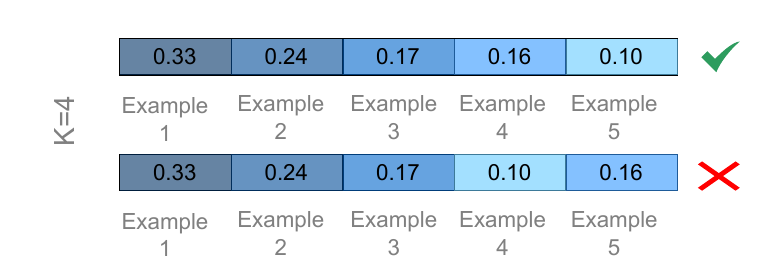}
            \caption{The top sample follows the examined positional bias ($K=4$) as the contributions of all the examples are sorted in decreasing order. The bottom sample does not follow the bias, as the fourth example breaks the monotonicity.}
        \label{fig:pos_type_ex_k4}
      \end{subfigure}
      
      \caption{For each of the examined positional bias types we illustrate 2 examples. One that follows the examined type of positional bias and one that does not. We note that the demonstrated examples are provided for purely illustrative purposes and do not depict any real data.}
      \label{fig:pos_bias_types_examples}
\end{figure*}

\begin{figure*}[t]
    \centering
    \includegraphics[width=\textwidth]{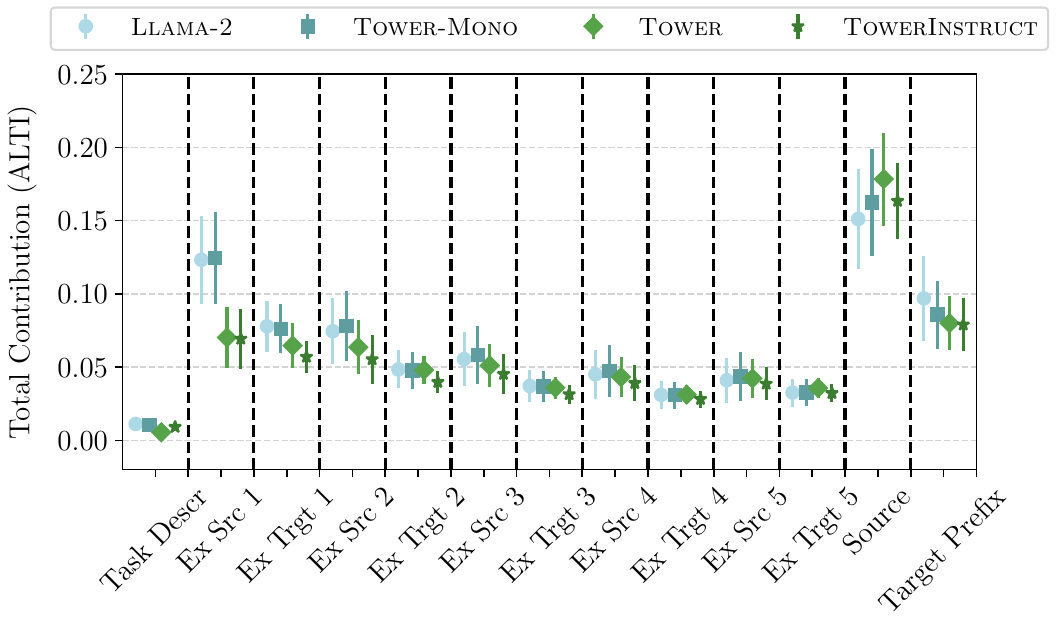}
    \captionsetup{width=\textwidth}
    \caption{Illustration of context's part-level contributions, when the task description is added. Translation direction: \textit{English to German}}
\label{fig:preamble_en_de}
\end{figure*}

\begin{figure*}[t]
    \centering
    \includegraphics[width=\textwidth]{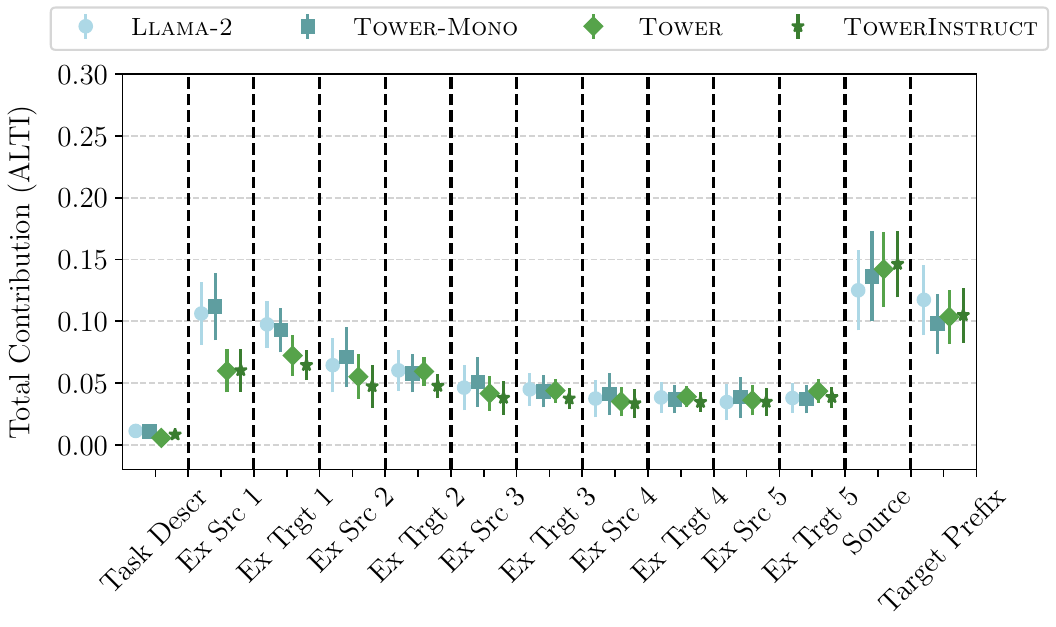}
    \captionsetup{width=\textwidth}
    \caption{Illustration of context's part-level contributions, when the task description is added. Translation direction: \textit{English to Russian}}
\label{fig:preamble_en_ru}
\end{figure*}

\begin{figure*}[t]
    \centering
    \includegraphics[width=\textwidth]{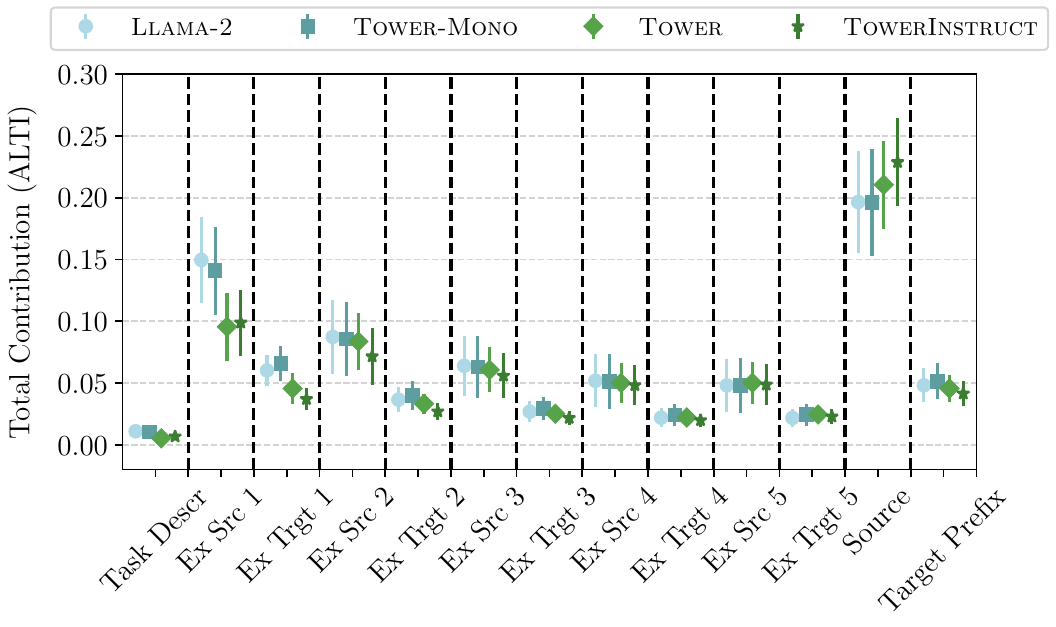}
    \captionsetup{width=\textwidth}
    \caption{Illustration of context's part-level contributions, when the task description is added. Translation direction: \textit{Russian to English}}
\label{fig:preamble_ru_en}
\end{figure*}

\begin{figure*}[t]
    \centering
    \begin{subfigure}[b]{0.49\linewidth}
        \centering
        \includegraphics[width=\linewidth]{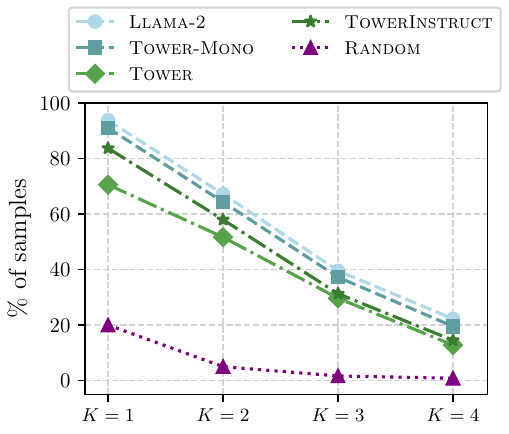}
        \caption{}
        \label{fig:pos_bias_analysis_en_de_rel}
    \end{subfigure}
    \begin{subfigure}[b]{0.49\linewidth}
        \centering
        \includegraphics[width=\linewidth]{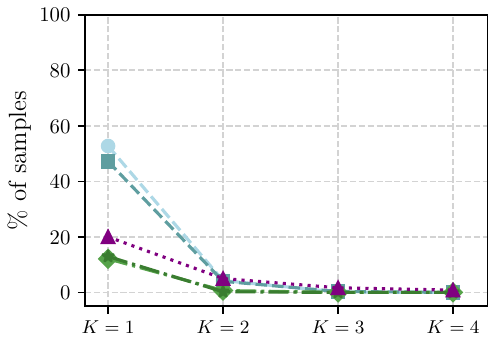}
        \caption{}
        \label{fig:pos_bias_break_en_de_rel}
    \end{subfigure}
    \caption{Proportion of \texttt{en-de} samples that follow positional bias, for different values of $K$, in the (a)~original and (b)~\texttt{replace-last-ex} settings.}
    \label{fig:pos_bias_analysis_and_break_en_de}
\end{figure*}

\begin{figure*}[t]
    \centering
    \begin{subfigure}[b]{0.49\linewidth}
        \centering
        \includegraphics[width=\linewidth]{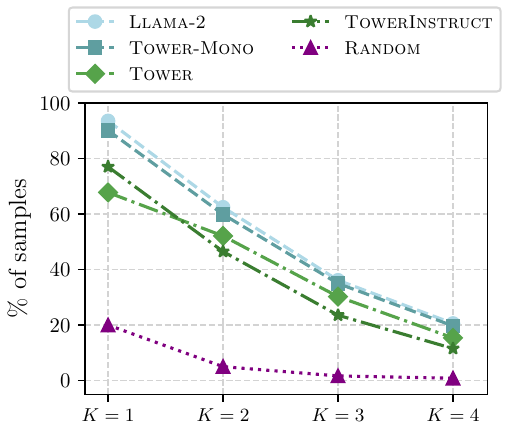}
        \caption{}
        \label{fig:pos_bias_analysis_ru_en_rel}
    \end{subfigure}
    \begin{subfigure}[b]{0.49\linewidth}
        \centering
        \includegraphics[width=\linewidth]{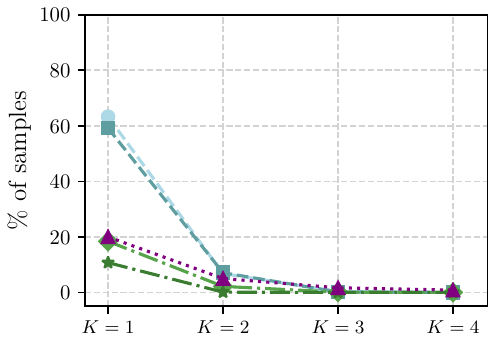}
        \caption{}
        \label{fig:pos_bias_break_ru_en_rel}
    \end{subfigure}
    \caption{Proportion of \texttt{ru-en} samples that follow positional bias, for different values of $K$, in the (a)~original and (b)~\texttt{replace-last-ex} settings.}
    \label{fig:pos_bias_analysis_and_break_ru_en}
\end{figure*}

\begin{figure*}[t]
    \centering
    \begin{subfigure}[b]{0.49\linewidth}
        \centering
        \includegraphics[width=\linewidth]{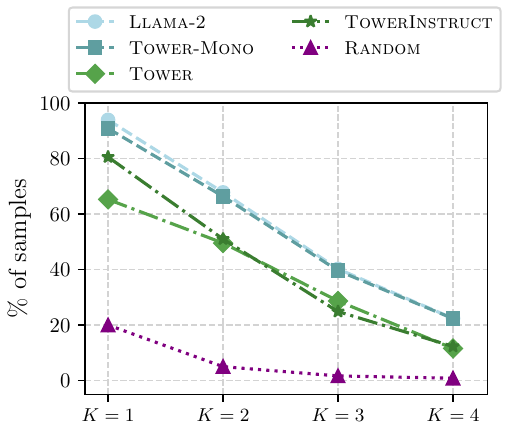}
        \caption{}
        \label{fig:pos_bias_analysis_en_ru_rel}
    \end{subfigure}
    \begin{subfigure}[b]{0.49\linewidth}
        \centering
        \includegraphics[width=\linewidth]{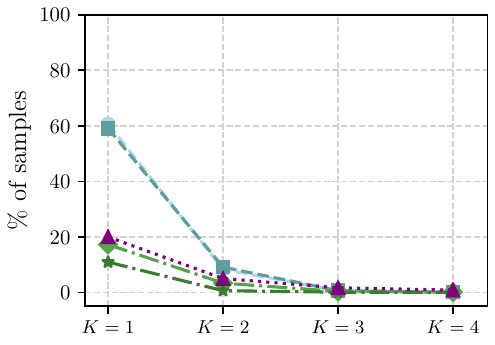}
        \caption{}
        \label{fig:pos_bias_break_en_ru_rel}
    \end{subfigure}
    \caption{Proportion of \texttt{en-ru} samples that follow positional bias, for different values of $K$, in the (a)~original and (b)~\texttt{replace-last-ex} settings.}
    \label{fig:pos_bias_analysis_and_break_en_ru}
\end{figure*}

\section{Top-level Analysis}
\label{app:top_level_analysis}
In the top-level analysis conducted in Section~\ref{sec1:high_level_analysis_segment_contributions}, we examined the contributions of individual parts of the context to the translated sequence and highlighted several findings. In addition, we provide results for the Russian to English and English to Russian language pairs (\S~\ref{app:high_level_russian}).
As supplementary material, we include an additional experiment (\S~\ref{app:reshuffling}) to enhance the validity of our findings, and we also present examples exhibiting anomalous part-level contributions (\S~\ref{app:anomalous_examples_high_level}) for completeness.

\subsection{Context's part-level contributions for additional language pairs}\label{app:high_level_russian}
In Figure~\ref{fig:pos_bias_dataset_level_ru_en_rel}, we show, for all the examined models, the total contribution of each context part to the translated sequence for Russian to English and English to Russian language pairs. We observe that results are largely similar with those presented in the main text for the German to English and English to German language pairs.

\subsection{Additional experiment by reshuffling the order of few-shot examples}\label{app:reshuffling}
 To ensure our findings hold against any potential, yet highly unlikely, content-related bias stemming from the position of the few-shot examples, we conduct a supplementary experiment. Put simply, we reshuffle the order of the few-shot examples for each sample and repeat the analysis. We report the results in Figures~\ref{fig:high_level_contr_reshuffle} and~\ref{fig:high_level_contr_reshuffle_russian} for German and Russian languages respectively. The top-level part-level contributions remain largely consistent with those presented in the main text. This result underscores the validity of the findings presented in Section~\ref{sec1:high_level_analysis_segment_contributions}.

\subsection{Examples with anomalous part-level contributions}\label{app:anomalous_examples_high_level}

In Figures~\ref{fig:hal_t_ex_575} and~\ref{fig:hal_t_ex_653}, we include some additional cases where the models hallucinate by copying one of the provided few-shot examples. We observe that in all cases the models exhibit anomalous  contributions and particularly the contribution of the source is minimal.  We also closely inspect similar cases in Appendix~\ref{app:generation_stages_anamalous_examples}, where we analyze the context dynamics across the generation stages and we discuss our findings.

\section{Positional Bias Analysis}
\label{app:pos_bias_analysis}

\subsection{Details on analysis setup and examples of positional bias types}
In the  analysis conducted in Section~\ref{subsec:positional_bias_analysis}, we assess the prevalence and the extent of the positional bias observed. Particularly, we examine whether the contributions of the first $K$ few-shot examples monotonically dominate the remaining $N-K$ examples. We consider different values of $K$ to represent the different types of positional bias. For instance, when $K=1$, the first few-shot example attains the highest level of contribution. In the case where $K=2$, the first two examples exhibit sorted contributions in a descending order and the remaining three have lower contributions than the first two, but they are not necessarily sorted in a descending order. Similarly, in the case where $K=3$, the first three few-shot examples exhibit  sorted contributions in a descending order and the remaining two have lower contributions than the first three, but they are not necessarily sorted in a descending order. Finally, when $K=4$, the few-shot examples exhibit globally monotonic contributions, indicating a strong positional bias across all examples. We visually illustrate examples of the aforementioned cases in Figure~\ref{fig:pos_bias_types_examples}.

\subsection{Additional plots} 

\paragraph{Is it all about position?} 
In Figures~\ref{fig:preamble_en_de},~\ref{fig:preamble_en_ru} and~\ref{fig:preamble_ru_en} 
we show the context's part-level contributions, when the task description is added for the English to German, English to Russian and Russian to English translation directions respectively. We notice that in all translation directions the task description receives significantly lower contribution compared to the examples and other parts of the context, suggesting that the positional bias is not merely a function of absolute position.

\paragraph{Can relevance to the test example break the bias?}
In Figures~\ref{fig:pos_bias_analysis_en_de_rel} and ~\ref{fig:pos_bias_break_en_de_rel}, we present the proportion of \texttt{en-de} samples that follow positional bias, for different values of $K$, in the original and \texttt{replace-last-example} settings respectively. We additionally provide the corresponding results for the Russian to English and  English to Russian translation directions in Figures~\ref{fig:pos_bias_analysis_and_break_ru_en} and~\ref{fig:pos_bias_analysis_and_break_en_ru} respectively.
In all settings examined, we observe that results are largely similar with those presented in Sections  \ref{subsec:positional_bias_analysis} and \ref{subsec:breaking_pos_bias}.

\section{Context Contributions across Generation Stages}
\label{app:generation_stages}
In Section \ref{sec:generation_stages}, we explored how context contributions evolve across different stages of the generation process for the \textsc{Tower} model. In the following part, we include additional plots examining how  context contributions evolve across the generation process for the rest of the models and language pairs examined. We additionally show examples of anomalous context contributions and other salient cases and we discuss the results.

\subsection{Additional plots}\label{app:generation_stages_plots}
\begin{figure*}[t]
    \centering
    \begin{subfigure}[t]{\textwidth}
        \centering
\includegraphics[width=\textwidth]{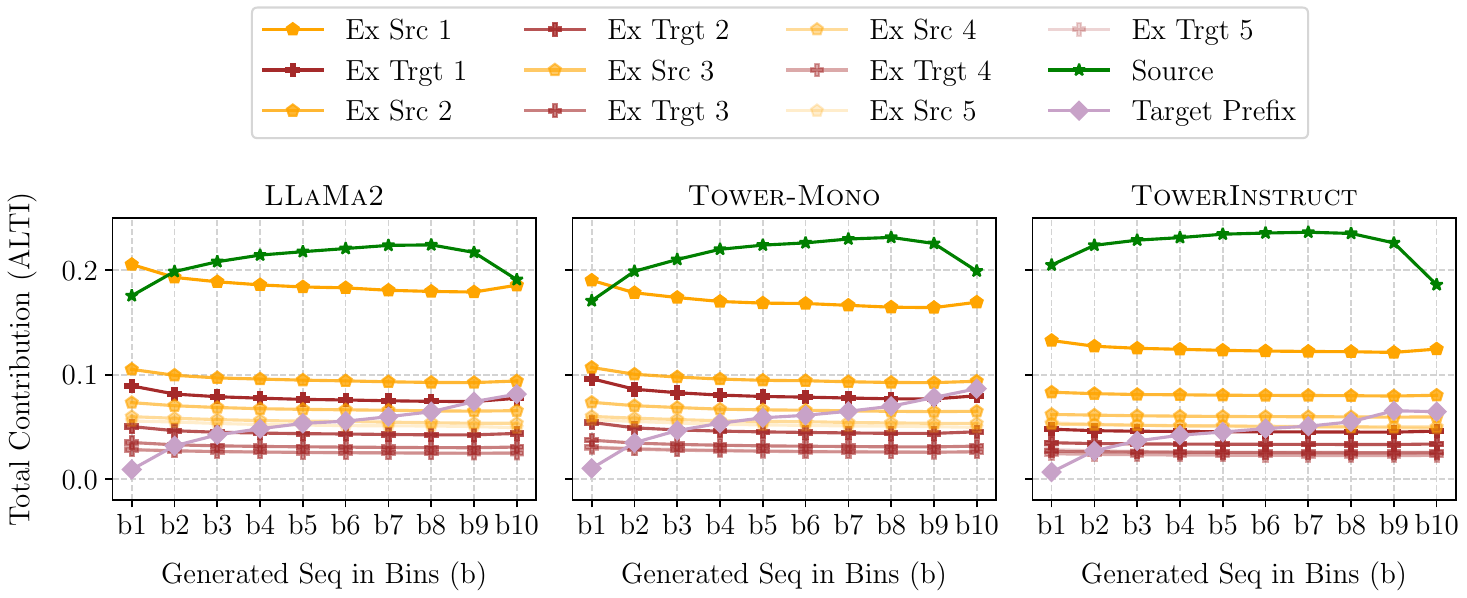}
        \caption{German to English}
    \end{subfigure}
    \vspace{1em} 
    \begin{subfigure}[t]{\textwidth}
        \centering
\includegraphics[width=\textwidth]{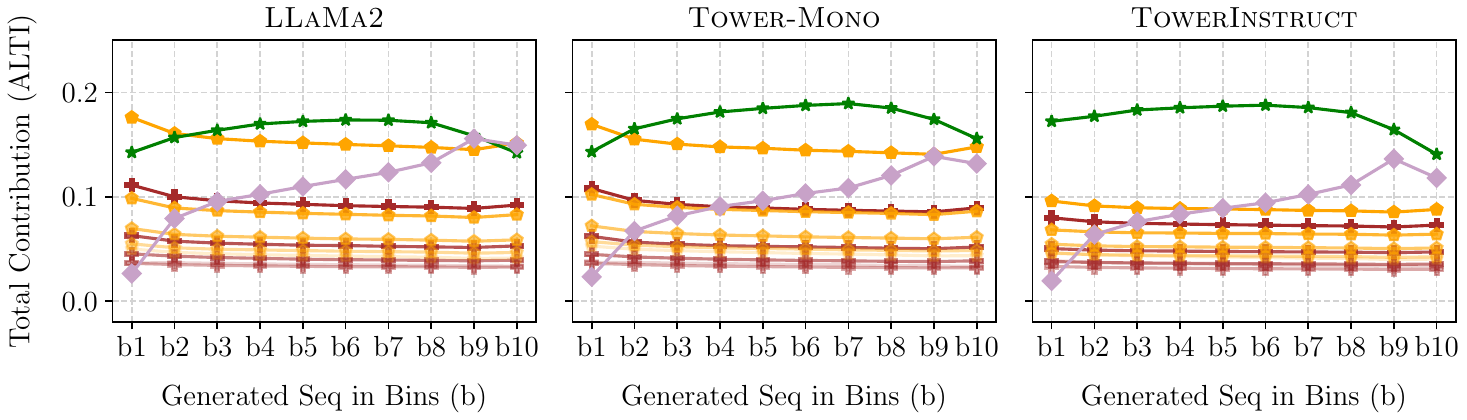}
        \caption{English to German}
    \end{subfigure}
    \caption{Illustration of how context contributions evolve across different generation stages, for the \textsc{Llama-2}, \textsc{Tower-Mono} and  \textsc{TowerInstruct} models. Each generated bin accounts for 10\% of the generated sequence.}
    \label{fig:dynamics_remaining}
\end{figure*}

\begin{figure*}[t]
    \centering
    \includegraphics[width=\textwidth]{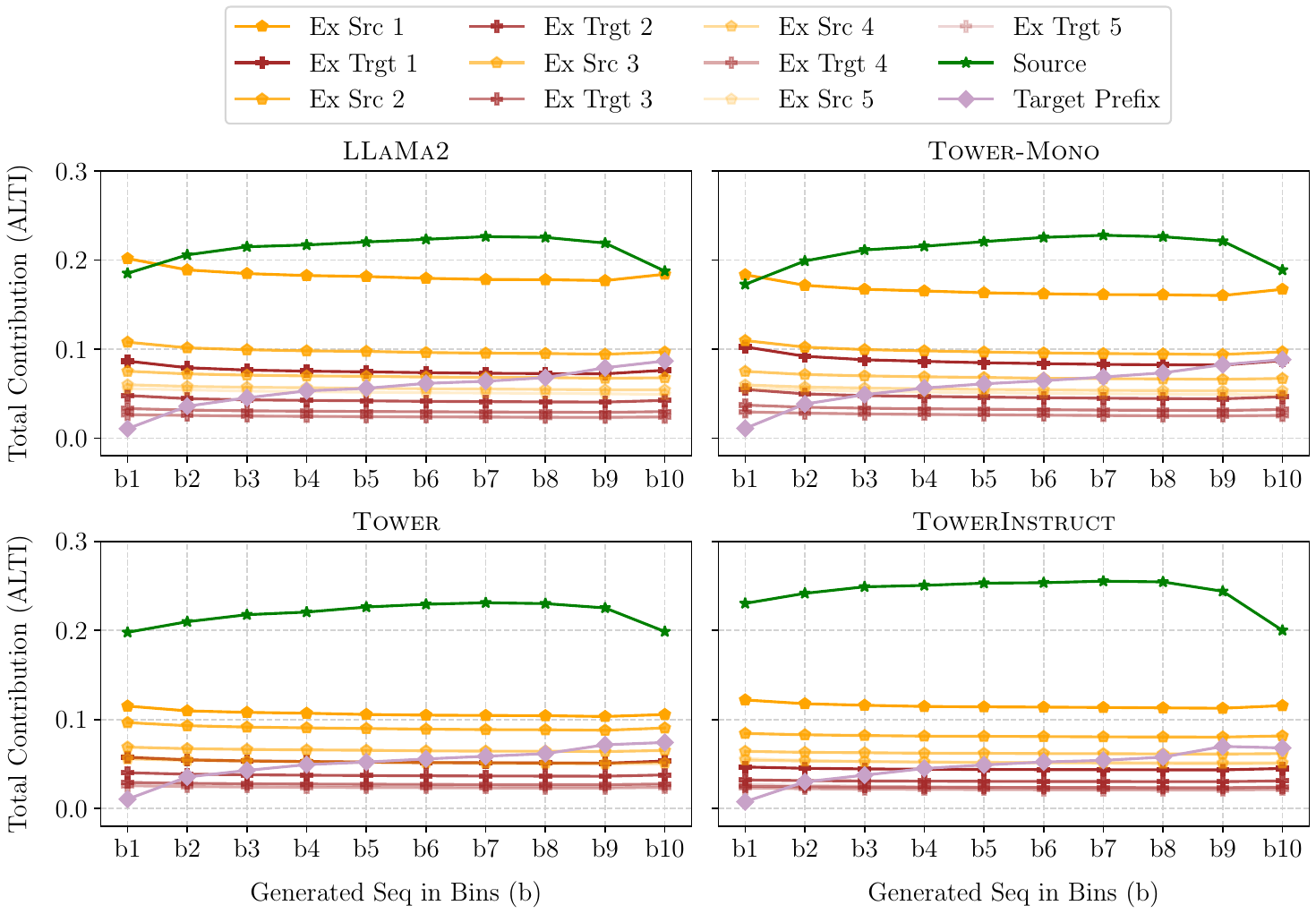}
    \captionsetup{width=\textwidth}
    \caption{Illustration of how context contributions evolve across different generation stages, for all the examined models. Each generated bin accounts for 10\% of the generated sequence. Translation direction: \textit{Russian to English}}
\label{fig:dynamics_ru_en}
\end{figure*}

\begin{figure*}[t]
    \centering
    \includegraphics[width=\textwidth]{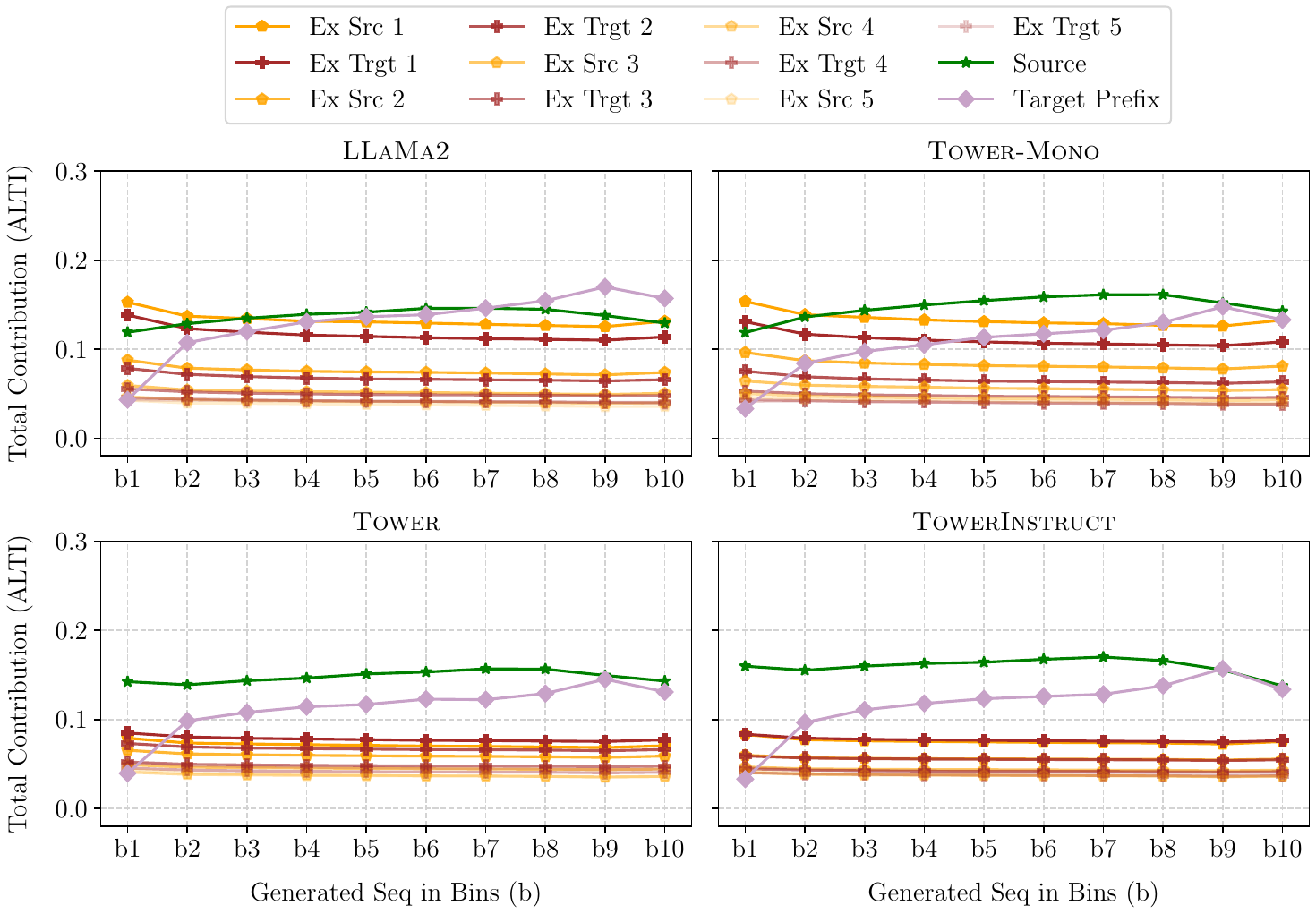}
    \captionsetup{width=\textwidth}
    \caption{Illustration of how context contributions evolve across different generation stages, for all the examined models. Each generated bin accounts for 10\% of the generated sequence. Translation direction: \textit{English to Russian}}
\label{fig:dynamics_en_ru}
\end{figure*}

In  Figure \ref{fig:dynamics_remaining}, we present how context contributions evolve across different generation stages for \textsc{Llama-2}, \textsc{Tower-Mono} and \textsc{TowerInstruct} models, for the de-en and en-de translation directions. For completeness, we provide in Figures~\ref{fig:dynamics_ru_en} and~\ref{fig:dynamics_en_ru} the corresponding plots for the ru-en and en-ru language pairs respectively.

\begin{table}[t]
\footnotesize
\centering
    \begin{tabular}{ccc}
        \toprule
        \textbf{Language Pair} & \textbf{Model} &\textbf{\# of hall.} \\
        \midrule
        En-De & \textsc{llama-2}&3\\
        En-De & \textsc{tower-mono}&4\\
        En-De & \textsc{tower}&1\\
        En-De & \textsc{towerinstruct}&1\\
        De-En & \textsc{llama-2}&2\\
        De-En & \textsc{tower-mono}&2\\
        De-En & \textsc{tower}&11\\
        De-En & \textsc{towerinstruct}&0\\
        En-Ru & \textsc{llama-2}&23\\
        En-Ru & \textsc{tower-mono}&4\\
        En-Ru & \textsc{tower}&10\\
        En-Ru & \textsc{towerinstruct}&1\\
        Ru-En & \textsc{llama-2}&1\\
        Ru-En & \textsc{tower-mono}&5\\
        Ru-En & \textsc{tower}&2\\
        Ru-En & \textsc{towerinstruct}&1\\
        \bottomrule
    \end{tabular}
    \caption{Number of fully detached hallucination cases by language pair and model.}
    \label{tab:app_hallucinations}
\end{table}

\subsection{Examples of anomalous context contributions and other salient cases} \label{app:generation_stages_anamalous_examples}

In Section~\ref{sec:generation_stages}, we highlighted the importance of anomalous source-part contributions as indicators of pathological translations. Here, we include more such examples as well as instances of other salient cases.

In Tables~\ref{tab:anomalous_example_569},~\ref{tab:anomalous_example_181} and~\ref{tab:anomalous_example_653},  we present 3 examples where one of the examined models hallucinates, exhibiting anomalous contributions. The example shown in Table~\ref{tab:anomalous_example_569} is particularly interesting, as both models in the beginning of the translation process exhibit low source contributions --- compared to the source-part contribution of the first example --- indicating that they primarily rely on the first example. However, as the translation progresses, the source contributions of the examined models follow completely opposite trends. \textsc{Tower} exhibits extremely anomalous contributions --- a steeply increasing contribution from the source-part of the first example and a decreasing one from the source --- producing in this way a hallucination, by copying the first example. In contrast, \textsc{Llama-2} produces a correct translation, with its contributions following the average case trends for German to English translation. Importantly, in all the provided examples, the models that produce a correct translation exhibit contribution trends that align with the average case trends we presented for German to English translation (see Figures~\ref{fig:dynamics_tower} and~\ref{fig:dynamics_remaining} for \textsc{Tower} and \textsc{Llama-2} respectively). 

Let's now turn to some other salient cases. In particular, we now turn to examples where the models do not produce any pathological translations~(see Tables~\ref{tab:non_hal_example_643} and~\ref{tab:non_hal_example_562}). Note that the models exhibit low source contributions in the early steps of the translation process (compared to the contributions of the few-shot examples) indicating a greater influence from the few-shot examples that are semantically similar. Then, as the translation progresses, they exhibit increased source contributions being very similar with the average case trends for German to English translation (see Figures~\ref{fig:dynamics_tower} and~\ref{fig:dynamics_remaining} for \textsc{Tower} and \textsc{Llama-2} respectively), indicating the reliance on the source to produce a correct translation.

\subsection{Details of Quantitative Analysis} \label{app:quantitative}
In Section~\ref{sec:generation_stages}, we examined whether anomalous context contributions can serve as indicators of hallucinations. Specifically we focused on how low source contributions, by conducting a quantitative analysis to assess the extent to which low-source contributions can be associated with "fully-detached" hallucinations. In this section, we provide further details regarding the annotation process.
\par For each model and language pair combination, we identify instances of "fully-detached" hallucinations by annotating the generated translations using the \textsc{Llama-3-70B-Instruct} model~\cite{dubey2024llama3herdmodels}, following the exact approach outlined by~\citealp{benkirane2024machinetranslationhallucinationdetection}.\footnote{We used the "Severity Ranking Prompt 2" as this was shown to be the optimal prompt for \textsc{Llama-3-70B-Instruct}.} Specifically, each instance is annotated into one of four categories: "No hallucination", "Small hallucination", "Partial hallucination", and "Full hallucination". Only instances classified as "Full hallucination" are considered "fully-detached" hallucinations in our analysis.
We report the number of full hallucinations for each of model and language pair combination in Table~\ref{tab:app_hallucinations}.

\section{AI Assistants}
We have used Github Copilot\footnote{https://github.com/features/copilot} during development of our research work.

\begin{table*}[t]
  \centering
  \footnotesize
  \begin{tabular}{lp{0.5\textwidth}l} 
    \cmidrule[\heavyrulewidth]{1-2}
      \centering
        \hlexampleone{\texttt{\textbf{E1}|SRC}} & Ich interessiere mich für das Objekt 08867 in Salzburg-Parsch &\,\, \multirow{13}{*}[2mm]{\includegraphics[width=6cm]{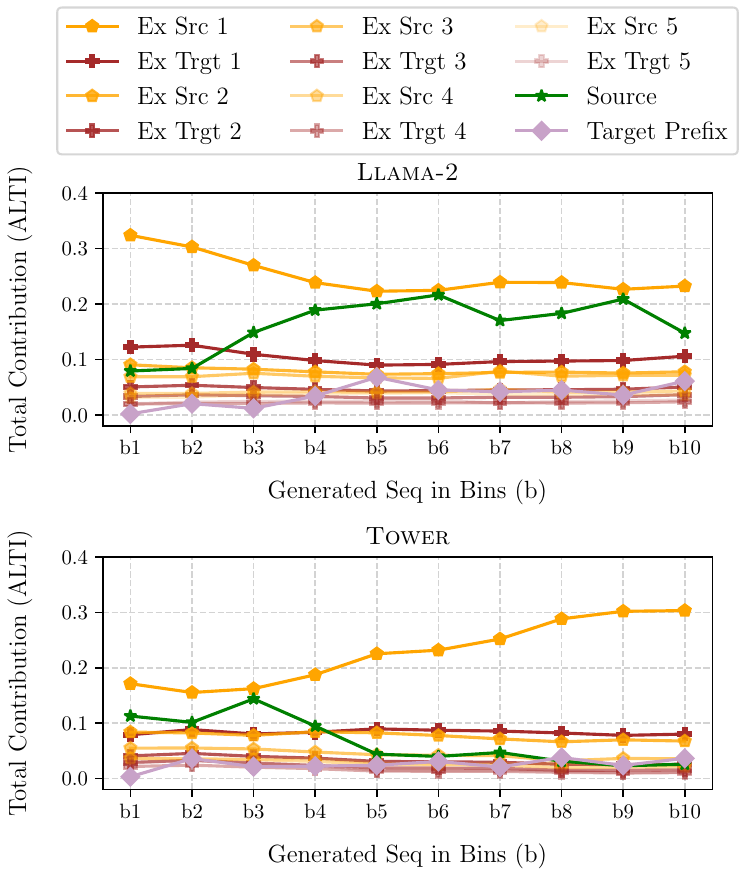}}\\
        \hlexampleonetgt{\texttt{\textbf{E1}|TGT}} & I am interested in the object 08867 in Salzburg-Parsch& \\\cdashlinelr{1-2}
        \hlexampleone{\texttt{\textbf{E2}|SRC}} & Ich interessiere mich für das Objekt 55057 in Salzburg-Itzling& \\
        \hlexampleonetgt{\texttt{\textbf{E2}|TGT}} & I am interested in the object 55057 in Salzburg-Itzling& \\\cdashlinelr{1-2}
        \hlexampleone{\texttt{\textbf{E3}|SRC}} &
        Ich interessiere mich für '2 bedrooms Apartment in Los Angeles.\\
        \hlexampleonetgt{\texttt{\textbf{E3}|TGT}} &
        I am interested in '2 bedrooms Apartment in Los Angeles.& \\\cdashlinelr{1-2}
        \hlexampleone{\texttt{\textbf{E4}|SRC}} &
        Ich interessiere mich für 'Apartment for rent in SAN DIEGO....'.& \\
        \hlexampleonetgt{\texttt{\textbf{E4}|TGT}} & 
        I am interested in 'Apartment for rent in SAN DIEGO....'.& \\\cdashlinelr{1-2}
        \hlexampleone{\texttt{\textbf{E5}|SRC}} & 
        Ich interessiere mich für das Objekt 33405 in Salzburg-Herrnau  & \\
        \hlexampleonetgt{\texttt{\textbf{E5}|TGT}} &
        I am interested in the object 33405 in Salzburg-Herrnau & \\\cdashlinelr{1-2}
        \hlsrc{\texttt{\textbf{SRC}}} & ich interessiere mich für den \#{PRS\_ORG}\# Stuhl. & \\\cdashlinelr{1-2}
        \multicolumn{2}{l}{\textcolor{black!70}{\textit{\textsc{Llama-2}}} \textcolor{darkgreen}{\ding{51}}}\\ 
        \hlmt{\texttt{\textbf{MT}}} & I am interested in the \#{PRS\_ORG}\# Chair.\\\cdashlinelr{1-2}
        \multicolumn{2}{l}{\textcolor{black!70}{\textit{\textsc{Tower}}} \textcolor{red}{\ding{55}}}\\ 
        \hlmt{\texttt{\textbf{MT}}} & I am interested in the object 08867 in Salzburg-Parsch\\\cmidrule[\heavyrulewidth]{1-2}

  \end{tabular}
  \caption{Illustration of an example exhibiting anomalous source contributions for \textsc{Tower} --- which hallucinates,  followed by \textsc{Llama-2}'s contributions, which performs normally.}
  \label{tab:anomalous_example_569}
\end{table*}

\begin{table*}[t]
  \centering
  \footnotesize
  \begin{tabular}{lp{0.5\textwidth}l} 
    \cmidrule[\heavyrulewidth]{1-2}
      \centering
        \hlexampleone{\texttt{\textbf{E1}|SRC}} &  Wie lange dauert es von Cefalù nach Taormina zu kommen? &\,\, \multirow{13}{*}[2mm]{\includegraphics[width=6cm]{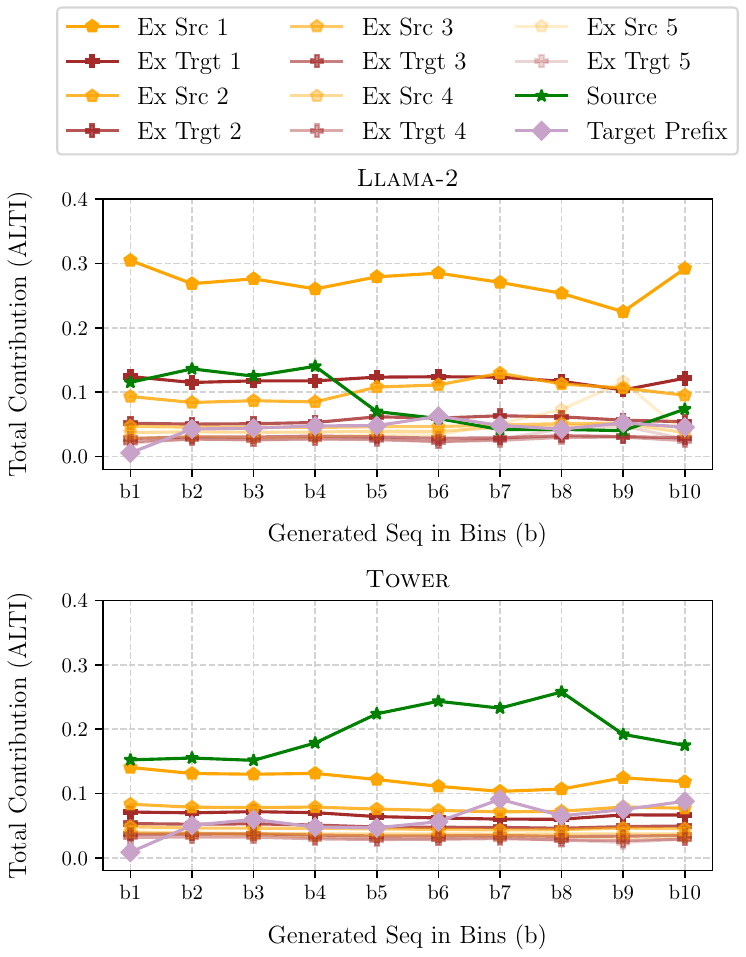}}\\
        \hlexampleonetgt{\texttt{\textbf{E1}|TGT}} & How long does it take to get from Cefalù to Taormina?& \\\cdashlinelr{1-2}
        \hlexampleone{\texttt{\textbf{E2}|SRC}} & Wie lange dauert es von Oslo nach Haugesund zu kommen?& \\
        \hlexampleonetgt{\texttt{\textbf{E2}|TGT}} & How long does it take to get from Oslo to Haugesund?& \\\cdashlinelr{1-2}
        \hlexampleone{\texttt{\textbf{E3}|SRC}} &
        Wie lange dauert es von Basel nach Montpellier zu kommen?\\
        \hlexampleonetgt{\texttt{\textbf{E3}|TGT}} &
        How long does it take to get from Basel to Montpellier?& \\\cdashlinelr{1-2}
        \hlexampleone{\texttt{\textbf{E4}|SRC}} &
        Wie lange dauert es von Flensburg nach Århus zu kommen?& \\
        \hlexampleonetgt{\texttt{\textbf{E4}|TGT}} & 
        How long does it take to get from Flensburg to Århus?& \\\cdashlinelr{1-2}
        \hlexampleone{\texttt{\textbf{E5}|SRC}} & 
        Wie lange dauert es von Oslo nach Hammerfest zu kommen?   & \\
        \hlexampleonetgt{\texttt{\textbf{E5}|TGT}} &
        How long does it take to get from Oslo to Hammerfest? & \\\cdashlinelr{1-2}
        \hlsrc{\texttt{\textbf{SRC}}} &wie lange dauert es die gelben zu bestellen mit und ohne armlehne?  & \\\cdashlinelr{1-2}
        \multicolumn{2}{l}{\textcolor{black!70}{\textit{\textsc{Llama-2}}} \textcolor{red}{\ding{55}}}\\ 
        \hlmt{\texttt{\textbf{MT}}} & How long does it take to get from Oslo to Hammerfest?\\\cdashlinelr{1-2}
        \multicolumn{2}{l}{\textcolor{black!70}{\textit{\textsc{Tower}}} \textcolor{darkgreen}{\ding{51}}}\\ 
        \hlmt{\texttt{\textbf{MT}}} & how long does it take to order the yellow with and without armrest?\\\cmidrule[\heavyrulewidth]{1-2}

  \end{tabular}
  \caption{Illustration of an example exhibiting anomalous source contribution for \textsc{Llama-2} --- which hallucinates,  followed by \textsc{Tower}'s contributions, which performs normally.}
  \label{tab:anomalous_example_181}
\end{table*}

\begin{table*}[t]
  \centering
  \footnotesize
  \begin{tabular}{lp{0.5\textwidth}l} 
    \cmidrule[\heavyrulewidth]{1-2}
      \centering
        \hlexampleone{\texttt{\textbf{E1}|SRC}} &  Wir wünschen Ihnen einen angenehmen Aufenthalt in Maribor. &\,\, \multirow{13}{*}[2mm]{\includegraphics[width=6cm]{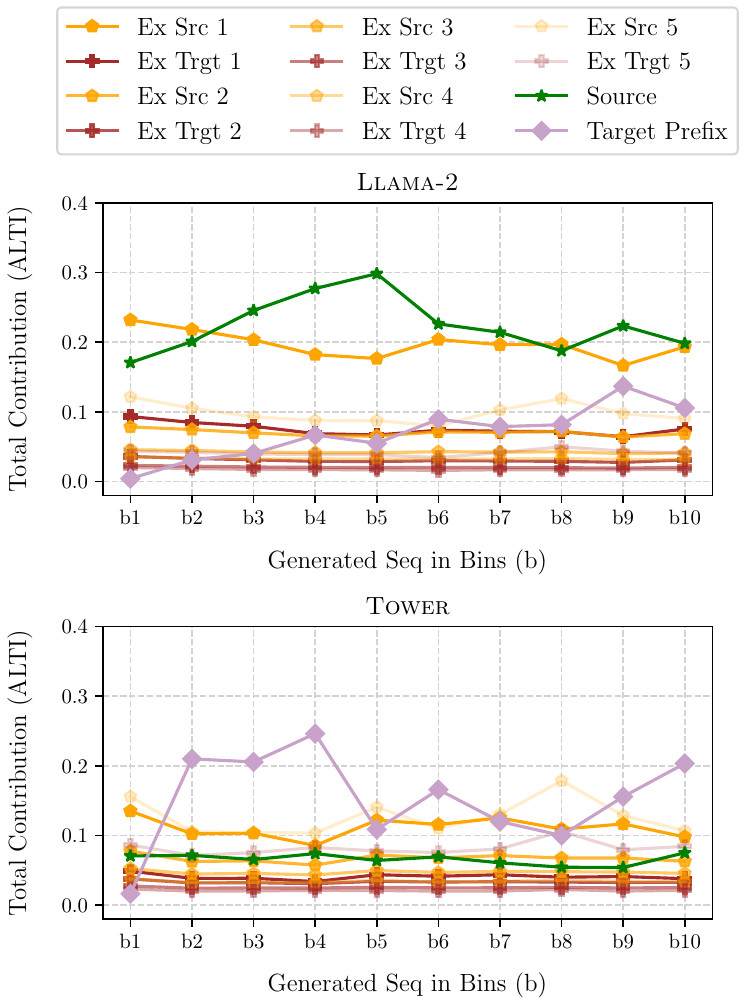}}\\
        \hlexampleonetgt{\texttt{\textbf{E1}|TGT}} & We wish you a pleasant stay in Maribor.& \\\cdashlinelr{1-2}
        \hlexampleone{\texttt{\textbf{E2}|SRC}} & Wir wünschen Ihnen einen angenehmen Aufenthalt in Olomouc.& \\
        \hlexampleonetgt{\texttt{\textbf{E2}|TGT}} & We wish you a pleasant stay in Olomouc.& \\\cdashlinelr{1-2}
        \hlexampleone{\texttt{\textbf{E3}|SRC}} &
        Wir wünschen Ihnen einen angenehmen Aufenthalt in Debrecen.\\
        \hlexampleonetgt{\texttt{\textbf{E3}|TGT}} &
        We wish you a pleasant stay in Debrecen.& \\\cdashlinelr{1-2}
        \hlexampleone{\texttt{\textbf{E4}|SRC}} &
        Wir wünschen Ihnen einen angenehmen Aufenthalt in Poznan.& \\
        \hlexampleonetgt{\texttt{\textbf{E4}|TGT}} & 
        We wish you a pleasant stay in Poznan.& \\\cdashlinelr{1-2}
        \hlexampleone{\texttt{\textbf{E5}|SRC}} & 
        Busbud hilft Ihnen, einen Bus von Lübeck nach Wismar zu finden.& \\
        \hlexampleonetgt{\texttt{\textbf{E5}|TGT}} &
        Busbud helps you find a bus from Lübeck to Wismar.& \\\cdashlinelr{1-2}
        \hlsrc{\texttt{\textbf{SRC}}} &Wir verraten Ihnen, wo Sie im Raum Lübeck doch noch einen Weihnachtsbraten herbekommen. & \\\cdashlinelr{1-2}
        \multicolumn{2}{l}{\textcolor{black!70}{\textit{\textsc{Llama-2}}} \textcolor{darkgreen}{\ding{51}}}\\ 
        \hlmt{\texttt{\textbf{MT}}} & We tell you where you can still get a Christmas roast in the Lübeck area.\\\cdashlinelr{1-2}
        \multicolumn{2}{l}{\textcolor{black!70}{\textit{\textsc{Tower}}} \textcolor{red}{\ding{55}}}\\ 
        \hlmt{\texttt{\textbf{MT}}} & Busbud helps you find a bus from Lübeck to Wismar.\\\cmidrule[\heavyrulewidth]{1-2}

  \end{tabular}
  \caption{Illustration of an example exhibiting anomalous source contribution for \textsc{Tower} --- which hallucinates,  followed by \textsc{Llama-2}'s contributions, which performs normally.}
  \label{tab:anomalous_example_653}
\end{table*}

\begin{table*}[t]
  \centering
  \footnotesize
  \begin{tabular}{lp{0.5\textwidth}l} 
    \cmidrule[\heavyrulewidth]{1-2}
      \centering
        \hlexampleone{\texttt{\textbf{E1}|SRC}} & Telefónica Deutschland hat den SABRE Award EMEA gewonnen. &\,\, \multirow{13}{*}[2mm]{\includegraphics[width=6cm]{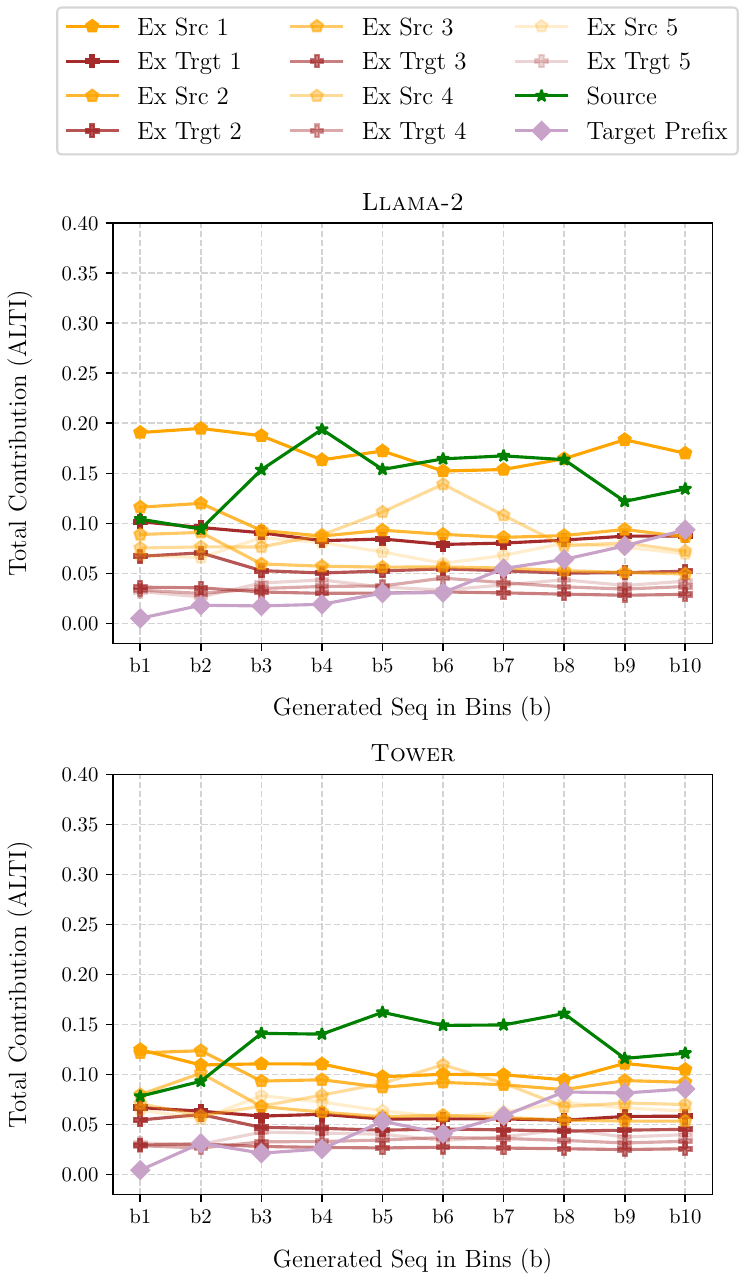}}\\
        \hlexampleonetgt{\texttt{\textbf{E1}|TGT}} & Telefónica Deutschland has won the SABRE Award EMEA.& \\\cdashlinelr{1-2}
        \hlexampleone{\texttt{\textbf{E2}|SRC}} & New York City (Bundesstaat New York, USA): Promenade im Central Park.& \\
        \hlexampleonetgt{\texttt{\textbf{E2}|TGT}} & New York city (New York State, USA): Promenade in Central Park.& \\\cdashlinelr{1-2}
        \hlexampleone{\texttt{\textbf{E3}|SRC}} &
       New York City FC oder New England Revolution\\
        \hlexampleonetgt{\texttt{\textbf{E3}|TGT}} &
        New York City FC or New England Revolution& \\\cdashlinelr{1-2}
        \hlexampleone{\texttt{\textbf{E4}|SRC}} &
        25.08 02:30 LA Galaxy - Los Angeles FC (Fußball,Major League Soccer)& \\
        \hlexampleonetgt{\texttt{\textbf{E4}|TGT}} & 
        25.08 02:30 LA Galaxy - Los Angeles FC (Calcio,Major League Soccer)& \\\cdashlinelr{1-2}
        \hlexampleone{\texttt{\textbf{E5}|SRC}} & 
        FC Schalke 04 hat 2 von den letzten 3 Spiele gegen VfL Wolfsburg gewonnen& \\
        \hlexampleonetgt{\texttt{\textbf{E5}|TGT}} &
        FC Schalke 04 has won 2 out of their last 3 matches against VfL Wolfsburg& \\\cdashlinelr{1-2}
        \hlsrc{\texttt{\textbf{SRC}}} &New York City FC hat zum ersten Mal den Titel in der Major League Soccer gewonnen. & \\\cdashlinelr{1-2}
        \multicolumn{2}{l}{\textcolor{black!70}{\textit{\textsc{Llama-2}}} \textcolor{darkgreen}{\ding{51}}}\\ 
        \hlmt{\texttt{\textbf{MT}}} & New York City FC has won the title in the Major League Soccer for the first time.\\\cdashlinelr{1-2}
        \multicolumn{2}{l}{\textcolor{black!70}{\textit{\textsc{Tower}}} \textcolor{darkgreen}{\ding{51}}}\\ 
        \hlmt{\texttt{\textbf{MT}}} &New York City FC has won the title in the Major League Soccer for the first time.\\\cmidrule[\heavyrulewidth]{1-2}

  \end{tabular}
  \caption{Illustration of an example where both \textsc{Llama-2} and \textsc{Tower} produce correct translations. We observe that their contributions follow the average case trends for German to English translation.}
  \label{tab:non_hal_example_643}
\end{table*}

\begin{table*}[t]
  \centering
  \footnotesize
  \begin{tabular}{lp{0.5\textwidth}l} 
    \cmidrule[\heavyrulewidth]{1-2}
      \centering
        \hlexampleone{\texttt{\textbf{E1}|SRC}} &Arminia Bielefeld - Union Berlin2. Bundesliga. &\,\, \multirow{13}{*}[2mm]{\includegraphics[width=6cm]{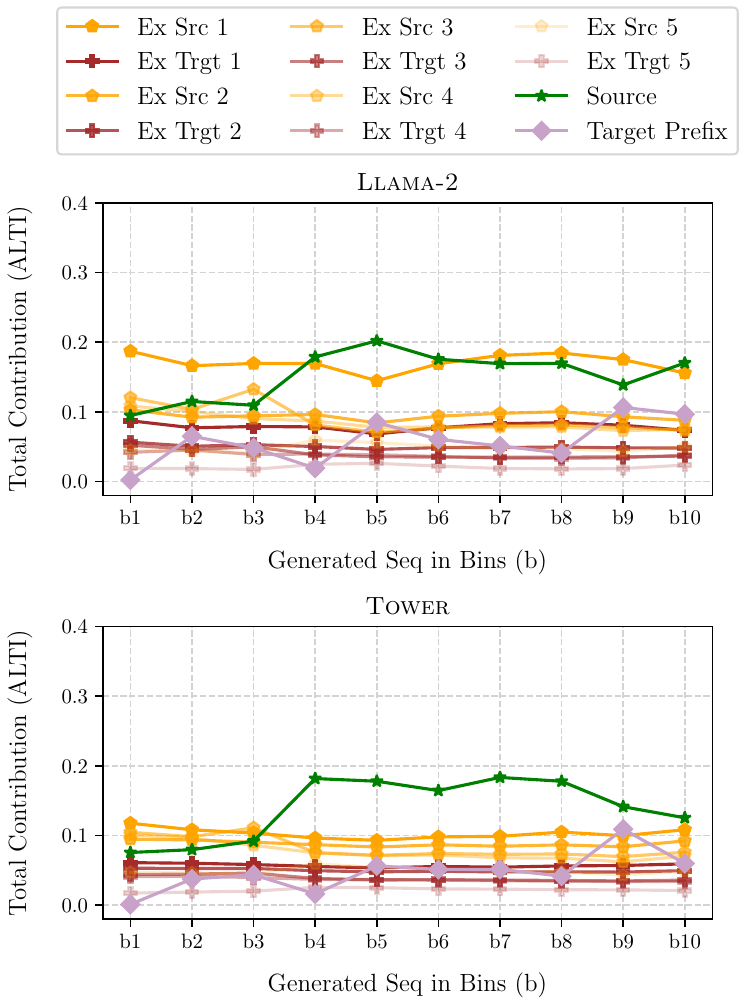}}\\
        \hlexampleonetgt{\texttt{\textbf{E1}|TGT}} & Arminia Bielefeld - Union Berlin2nd Bundesliga.& \\\cdashlinelr{1-2}
        \hlexampleone{\texttt{\textbf{E2}|SRC}} &Hertha BSC: Gewinner der 2. Bundesliga 2010/2011& \\
        \hlexampleonetgt{\texttt{\textbf{E2}|TGT}} &Hertha BSC: 2. Bundesliga winners 2010/2011& \\\cdashlinelr{1-2}
        \hlexampleone{\texttt{\textbf{E3}|SRC}} &
       Samstag, 9. März 2019 SV Darmstadt 98 Holstein Kiel\\
        \hlexampleonetgt{\texttt{\textbf{E3}|TGT}} &
        Saturday, 9 March 2019 SV Darmstadt 98 Holstein Kiel& \\\cdashlinelr{1-2}
        \hlexampleone{\texttt{\textbf{E4}|SRC}} &
        Darmstadt Reisen von Saarbrücken nach Darmstadt in 4 stunden und 59 minuten& \\
        \hlexampleonetgt{\texttt{\textbf{E4}|TGT}} & 
        Darmstadt Travel from Saarbrücken to Darmstadt in 4 hours and 59 minutes& \\\cdashlinelr{1-2}
        \hlexampleone{\texttt{\textbf{E5}|SRC}} & 
        Das Wasser darf nicht heißer als 60 °C sein.& \\
        \hlexampleonetgt{\texttt{\textbf{E5}|TGT}} &
        The water must not be hotter than 60 °C.& \\\cdashlinelr{1-2}
        \hlsrc{\texttt{\textbf{SRC}}} &Darmstadt 98 darf von der Rückkehr in die Fußball-Bundesliga träumen. & \\\cdashlinelr{1-2}
        \multicolumn{2}{l}{\textcolor{black!70}{\textit{\textsc{Llama-2}}} \textcolor{darkgreen}{\ding{51}}}\\ 
        \hlmt{\texttt{\textbf{MT}}} & Darmstadt 98 can dream of returning to the Bundesliga.\\\cdashlinelr{1-2}
        \multicolumn{2}{l}{\textcolor{black!70}{\textit{\textsc{Tower}}} \textcolor{darkgreen}{\ding{51}}}\\ 
        \hlmt{\texttt{\textbf{MT}}} &Darmstadt 98 can dream of a return to the Bundesliga.\\\cmidrule[\heavyrulewidth]{1-2}

  \end{tabular}
  \caption{Illustration of an example where both \textsc{Llama-2} and \textsc{Tower} produce correct translations. We observe that their contributions follow the average case trends for German to English translation.}
  \label{tab:non_hal_example_562}
\end{table*}

\end{document}